\documentclass[10pt,twocolumn,letterpaper]{article}

\usepackage[]{cvpr}  
\usepackage[ruled,linesnumbered]{algorithm2e}

\usepackage{booktabs}       % professional-quality tables
\usepackage{amsfonts}       % blackboard math symbols
\usepackage{nicefrac}       % compact symbols for 1/2, etc.
\usepackage{microtype}      % microtypography
\usepackage{xcolor}         % colors
\usepackage{amsmath}
\usepackage{graphicx}
\usepackage{multirow}
\usepackage{booktabs}
\usepackage{wrapfig}
\usepackage{enumitem}

\usepackage[ruled,linesnumbered]{algorithm2e}
\usepackage{algorithmic}
\usepackage{multirow}
\usepackage{booktabs}
\usepackage{wrapfig}
\usepackage{enumitem}

\definecolor{cvprblue}{rgb}{0.21,0.49,0.74}
\usepackage[pagebackref,breaklinks,colorlinks,citecolor=cvprblue]{hyperref}
\usepackage{graphicx}

\def\confName{CVPR}
\def\confYear{2024}

\begin{document}

\title{The Devil is in the Edges: Monocular Depth Estimation with Edge-aware Consistency Fusion}

\author{Pengzhi Li, \ \ Yikang Ding, \ \ Haohan Wang,\ \ Chengshuai Tang,\ \ Zhiheng Li$^{{\dag}}$\\
$^1$Tsinghua Shenzhen International Graduate School, Tsinghua University \ \
}

\maketitle

\begin{abstract}
This paper presents a novel monocular depth estimation method, named ECFNet, for estimating high-quality monocular depth with clear edges and valid overall structure from a single RGB image.
We make a thorough inquiry about the key factor that affects the edge depth estimation of the MDE networks, and come to a ratiocination that the edge information itself plays a critical role in predicting depth details.
Driven by this analysis, we propose to explicitly employ the image edges as input for ECFNet and fuse the initial depths from different sources to produce the final depth.
Specifically, ECFNet first uses a hybrid edge detection strategy to get the edge map and edge-highlighted image from the input image, and then leverages a pre-trained MDE network to infer the initial depths of the aforementioned three images.
After that, ECFNet utilizes a layered fusion module (LFM) to fuse the initial depth, which will be further updated by a depth consistency module (DCM) to form the final estimation.
Extensive experimental results on public datasets and ablation studies indicate that our method achieves state-of-the-art performance. Project page: \color{magenta}{https://zrealli.github.io/edgedepth/}.
\end{abstract}

\let\thefootnote\relax\footnotetext{
$^\dag$Corresponding author.
}

\begin{figure*}[t]
	\centering
	\includegraphics[width=\linewidth]{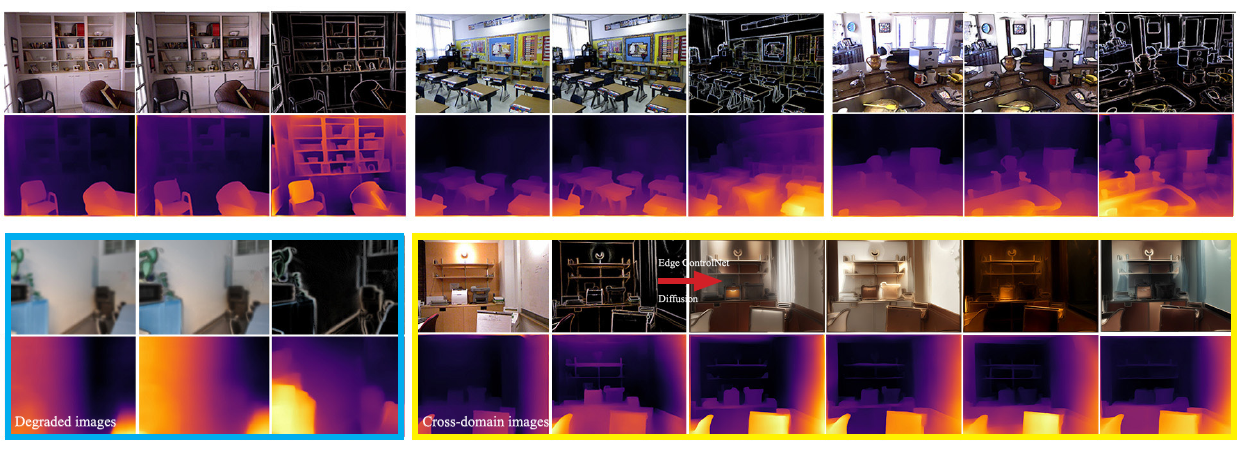}
	\caption{Depth visualizations on NYU-v2~\cite{nyu2012indoor}. For each triplet in the firsr row, we showcase the RGB image, the edge-highlighted map, the edge map, and their corresponding depth maps predicted by pre-trained DPT~\cite{Ranftl2021}. The edge maps are obtained using the hybrid edge detection strategy. In the second row, we show more edge observations in degraded images or cross-domain applications.}
	\label{fig:edge_nonedge}
	 % \vspace{-1em}
\end{figure*}

\section{Introduction}\label{sec:intro}

Monocular depth estimation (MDE) is a classical and fundamental computer vision task with a wide range of applications in autonomous driving, the metaverse, and robotics.

Without the depth data from depth sensors or geometric constraints from the posed multi-view images, it is tricky to recover the 3D structure of the observed scene from a single image. 
Recently, learning-based methods~\cite{bhat2021adabins,laina2016deeper,yin2021learning,Ranftl2022,wu2022toward} demonstrate great potential in this task by using networks, (\textit{e.g.}, CNN~\cite{Ranftl2022} or transformers~\cite{Ranftl2021}) to map the RGB colors to depth values.
Although various network architectures~\cite{johnston2020self,poggi2020uncertainty,fu2018deep}, loss functions~\cite{yin2021virtual,Xian_2020_CVPR}, and training strategies~\cite{depthattentionmodel,godard2017unsupervised,wong2019bilateral} have been proposed to improve the estimated depth, existing methods still suffer from missing details and over-smooth estimations in the final depth.
To handle this problem, several works exploit the edge information~\cite{yang2018unsupervised} and image segmentation~\cite{zhu2020edge} to enhance the depth details.
Although these methods have made progress in predicting depth maps with details, the estimated depth in edges is still struggling, leading to over-smooth final predictions.
Estimating high-quality depth maps with accurate edges and details remains a significant obstacle for MDE networks.

This paper aims to reduce the ambiguity in the important edges and detailed regions, as well as acquire finer overall depth predictions.
We ask: \textit{what has the greatest impact on edge depth}?
We perform various experiments and arrive at a conjecture: \textit{the edge itself hides the most critical information}.
The following analysis demonstrates our proposal from several perspectives.

(\textbf{\romannumeral1}) By inspecting the predicted depths of the existing MDE networks~\cite{Ranftl2021,Ranftl2022,yin2021learning}, we find that they can obtain competent results in obvious and large edge regions, (\textit{e.g.}, edges of large foreground objects), but their predictions in low-contrast edges, long distance, and small edges are very fuzzy and inaccurate as shown in Fig.~\ref{fig:edge_nonedge}.
We analyze and attribute this phenomenon to that networks do not capture all the edge information as the heavily-used convolution layers, downsampling and upsampling operations would smooth out small natural edge areas.
To validate this ratiocination, we test the same pretrained MDE network~\cite{Ranftl2021} with different inputs: the original images, the corresponding edge maps, and the images with highlighted edges (by deleting edge pixels).
As shown in Fig.~\ref{fig:edge_nonedge}, the edge maps and edge-highlighted images obtain clearer edges (please ignore the structural distortion in the depth maps). The aforementioned results indicate that the edge itself plays a critical role in producing fine details and edges.

(\textbf{\romannumeral2})
Another phenomenon that supports our conjecture is the performance drop of the degraded and cross-domain images. Some existing works~\cite{hu2019visualization,dijk2019neural} believe that MDE networks rely on geometric cues, occlusion boundaries, and textures to produce depth details. However, they cannot explain the significant performance gap between the normal and degraded images, as these factors would hardly be affected by the slight noise. 
Therefore, we attribute the performance drop to the disturbance of the edge information, which is more than sensitive to noise or blur.
As illustrated in Fig.~\ref{fig:edge_nonedge}(blue box), we compare the predicted depths on the degraded images, the corresponding edge maps, and the edge-highlighted images,
the latter two inputs turn out to be robust to the image degradation, which suggests the key to producing clearer edges is to preserve and utilize the edge information.
Furthermore, as shown in the yellow box, we utilize ControlNet's~\cite{zhang2023adding} edge control conditions and stable diffusion models~\cite{rombach2022high} to generate a series of images with diverse styles from the edge map. Despite the variations in texture and color information among these images, they consistently produce nearly identical depth maps. We attribute this phenomenon to the consistency in their edge structure information. Please see more details in the supplementary material.
Driven by this analysis, we propose to dig into the edge information and use it to produce high-quality depth maps with accurate edges and details.
Specifically, we present an Edge-aware Consistency Fusion network (ECFNet) which mainly consists of two parts: Layered Fusion Module (LFM) and Depth Consistency Module (DCM).
LFM plays a role in acquiring clear edges by fusing the initial depth maps from the edge maps, the edge-highlighted images, and the original depth maps.
The key idea of LFM derives from the fact that the edge itself helps predict sharper depth edges (see Fig.~\ref{fig:edge_nonedge}). As the quality of the detected edges would affect the fused depth, we develop a hybrid edge detection strategy to get high-quality edges, which incorporates the classical and learning-based edge detection methods.
However, as the edge maps contain no texture and shadow cues, the corresponding predicted depth maps suffer from the wrong spatial structure and inaccurate scale. What's more, even the same fixed MDE network can not guarantee the predictions from different inputs, (\textit{i.e.}, the original image, the corresponding edge map, and the edge-highlighted image) have the same depth range and distribution.
The aforementioned two problems lead to a thorny phenomenon: the fused depth maps from LFM have clear edges but inaccurate overall structure.
To handle this problem, we propose DCM to learn the local depth residual between the fused depth and the initial depth. The learned residual will be further used to update the fused depth.
The goal of DCM is to maintain the high-frequency details and reduce the structural errors in the fused depth, which is critical to get the high-quality final depth.

We conduct various experiments to compare ECFNet with other methods and to validate the effectiveness of different modules. Extensive experimental results indicate that ECFNet achieves state-of-the-art performance compared to related methods, especially in producing accurate edge depth.
We also illustrate that ECFNet achieves significantly superior performance to the existing \textit{SOTA} methods when faced with degraded images. 

Our contributions can be summarized as follows:

\begin{itemize}[noitemsep]
	\item We explore and validate that edge information plays a critical role in producing high-quality depth maps with clear edges and details, providing new insights for cross-domain downstream tasks utilizing clear edge depth maps (see details in supplementary material).
	\item We present a novel edge-aware consistency fusion network (ECFNet) to predict accurate monocular depth, which consists of a layered fusion module and a depth consistency module.
	\item Experimental results on multiple datasets indicate that our method achieves state-of-the-art performance.
    % \item 

\end{itemize}

\begin{figure*}[t]
	\centering

	\includegraphics[width=0.95\linewidth]{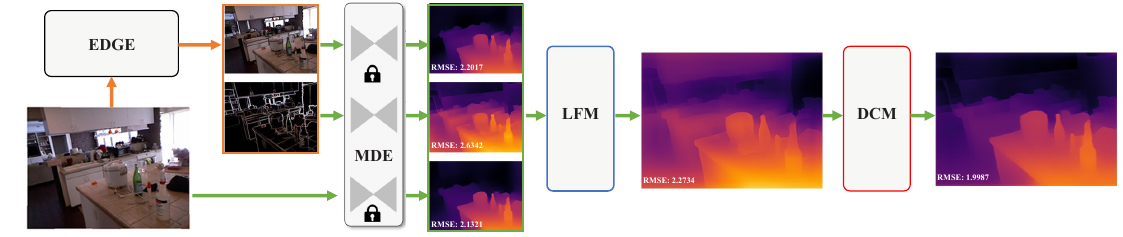}
  % \vspace{-0.7em}
	\caption{Pipeline of ECFNet. Given an image, ECFNet first extracts the edge map and computes the edge-highlighted image by removing edge pixels. These three images (including the original image) are fed into a frozen MDE network to predict initial depth maps as well. Subsequently, the initial depths are fused using LFM. Finally, the DCM is used to reduce the errors in fused depth from LFM and improve the depth consistency between the final depth and initial depth.}
	\label{fig:pipeline}
	% \vspace{-0.5em}
\end{figure*}

\section{Related Works}

\paragraph*{Monocular depth estimation} 
Monocular depth estimation aims to predict depth from a single RGB image. The existing works usually learn a non-linear mapping from pixels to depth values through deep neural networks~\cite{walia2022gated2gated,eigen2014depth,bhat2021adabins,laina2016deeper,yin2021learning,Ranftl2022,wu2022toward}. Remarkable progress in MDE has been witnessed in recent years, such as the large-scale depth datasets~\cite{chen2020oasis,kim2018deep,niklaus20193d,li2018megadepth}, the novel neural architecture designs~\cite{johnston2020self,poggi2020uncertainty}, the various loss functions~\cite{yin2021virtual,Xian_2020_CVPR,yin2021learning,lossrebalancing}, and the efficient training strategies~\cite{Ranftl2021,Ranftl2022,depthattentionmodel,godard2017unsupervised,wong2019bilateral,li2020unsupervised,li2023efficient}.

Nevertheless, existing MDE models~\cite{Ranftl2021,yin2021learning,Ranftl2022,Xian_2020_CVPR} often fail to predict depth maps with abundant high-frequency details, especially when encountered with degraded images. On the contrary, our approach enhances these details by incorporating the depth map generated from the edge map, and thus is able to predict clear and continuous edges in the depth map even in the degraded images.

\paragraph*{Edges in depth estimation}

Some recent works notice that the edge information would be helpful in training depth estimation networks~\cite{zhu2020edge,talker2022mind,wang2020sdc,yang2018unsupervised,ramamonjisoa2020predicting,qiu2020pixel}.
Zhu \textit{et al.}~\cite{zhu2020edge} propose to regularize the depth edges in the loss function where the edges come from the boundaries of the segmentation results.
Talker \textit{et al.}~\cite{talker2022mind} introduce a depth edges loss to enforce sharp edges in the estimated depth maps.
Several other works~\cite{ummenhofer2017demon, Xian_2020_CVPR,wang2019web} attempt to design gradient-based loss functions to improve the details.
Xian \textit{et al.}~\cite{Xian_2020_CVPR} develop an edge-guided sampling strategy to form a pair-wise ranking loss for enhancing the edge depth.
These methods mainly focus on how to utilize the edge information as an optimization constraint, \textit{i.e.}, use the edges in the loss function.
Our approach instead proposes to explicitly use the edge maps as input and fuse the estimated depth from different sources to improve the final depth.

\paragraph*{Depth refinement} 

Miangoleh \textit{et al.}~\cite{Miangoleh2021Boosting} propose a depth refinement method to merge depth from different resolutions and patches via exploiting the effect of the receptive field of the network on high-frequency details prediction.
However, this method requires high-quality input images, and its performance would significantly degrade when the input images contain slight noise. 
Similarly, Dai \textit{et al.}~\cite{dai2022multi} introduce another merging-based method, which is still sensitive to the image noise as it attempts to merge high-frequency details from high-resolution depth maps into low-resolution depth map images.
Unlike previous methods that rely on high-resolution input images to reserve the details in estimated depth, we focus on the edge information itself for producing high-quality edge depth.

\begin{figure}[t]
	\centering
	\includegraphics[width=0.99\linewidth]{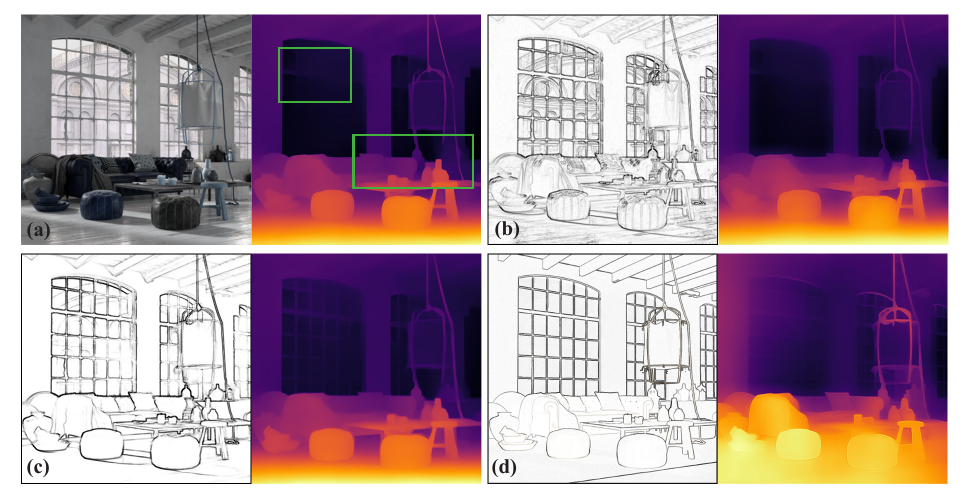}
	\caption{Comparison of different edge maps and their corresponding depth maps. (a) the RGB image, (b) the Sobel edge map~\cite{sobel1983accuracy}, (c) our edge map, and (d) the ground truth edge map. As the quality of the edge map improves, the corresponding depth map can capture more depth details (ignoring structural distortion).}
	\label{fig:edges}
	% \vspace{-1em}
\end{figure}

\section{Method}
In this section, we demonstrate our proposed method in detail. The overall pipeline of ECFNet is introduced in Sec.~\ref{sec:method_overview}. The proposed hybrid edge detection strategy is illustrated in Sec.~\ref{sec:method_edge}. The core parts of ECFNet (\textit{i.e.}, LFM and DCM) are introduced in Sec.~\ref{sec:method_lfm} and Sec.~\ref{sec:method_dcm} correspondingly.

\subsection{Overview}\label{sec:method_overview}

As analyzed in Sec.~\ref{sec:intro}, using the edge information is critical to predicting high-quality depth maps with accurate edges.
To this end, we introduce an edge-aware consistency fusion network, named ECFNet, to explicitly fuse the depth of edge maps with the corresponding depth of RGB images.
The data flow of ECFNet is shown in Fig.~\ref{fig:pipeline}. Given a single RGB image $I$, ECFNet first extracts the edge map $I_{e}$ via using the hybrid edge detection strategy (Sec.~\ref{sec:method_edge}). The corresponding edge-highlighted RGB image $I_{eh}$ is obtained by deleting the edge pixels in $I$.
After that, $I$, $I_e$, and $I_{eh}$ are used to infer initial depth $D$, $D_e$ and $D_{eh}$ using the base MDE model, which can be arbitrary pre-trained existing MDE networks.
As aforementioned, the $D_e$ has clear edges but lacks reasonable spatial structure due to the absence of texture and shadow in $I_e$.
On the contrary, the edges in $D$ and $D_{eh}$ are not clear enough, but their overall structure is more accurate.
As a result, the $D$, $D_e$, and $D_{eh}$ are fused by LFM to integrate their advantages.
% The LFM is implemented with shallow convolution layers, and the detailed network architecture can be found in supplementary materials.
The fused depth $D_{fuse}$ in LFM is not flawless. To fix the errors introduced by wrong spatial structure, inconsistent depth range, and depth distribution between the initial depth (\textit{i.e.}, $D$, $D_e$ and $D_{eh}$) and the $D_{fuse}$, ECFNet utilizes DCM to update the $D_{fuse}$.
Consequently, DCM maintains the high-frequency details and recovers the fine structure in the final produced depth.

\begin{figure}[t]
 
  \begin{center}
	\includegraphics[width=0.95\linewidth]{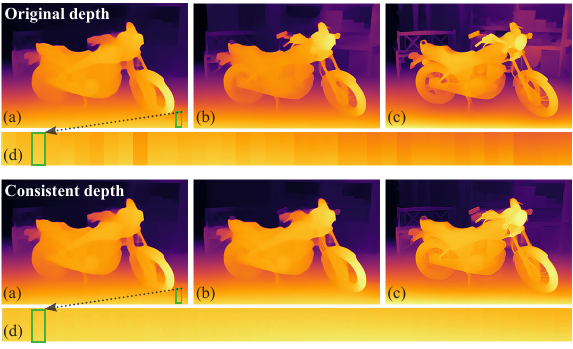}
	\caption{Visualized depth results before and after DCM. (a)-(c) represent the depth maps with different resolutions, while (d) displays depth slices of the green regions in the sample sequence. DCM helps significantly reduce the inconsistency between the input depth maps.}
	\label{fig:dcm}
  \end{center}
    % \vspace{-1em}
\end{figure}

\subsection{Edge Detection Strategy}\label{sec:method_edge}

% 图说明边缘越好深度越好。sobel缺点，learning缺点。我们怎么做
As shown in Fig. ~\ref{fig:edges}, improving the quality of edge maps could enrich the details in the depth map. However, it is non-trivial to obtain high-quality edge maps.
% As illustrated in Fig.~\ref{fig:edges_com},
Conventional edge detection algorithms, \textit{e.g.}, Sobel operator~\cite{sobel1983accuracy}, tend to produce unwanted artifacts and details, such as irregular texture edges, consequently leading to a blurred depth map. Learning-based edge detection methods~\cite{he2019bi,Pu_2022_CVPR,su2021pixel} could generate cleaner edges, but we experimentally find that the edges predicted are always a few pixels away from the accurate position. Hence in this work, we fuse the learning-based BDCN edges~\cite{he2019bi} with Sobel edges~\cite{sobel1983accuracy} by calculating their geometric mean for the sake of clean and sharp edges.

Furthermore, inspired by BMD~\cite{Miangoleh2021Boosting}, we increase the input resolution appropriately to enhance the accuracy of the estimated edge map. Concretely, we first divide the image into nine uniform patches, and upsample the patches to high-resolution ones. Then these patches are fed into the BDCN~\cite{he2019bi} network to extract edge maps, which will be fused with the Sobel~\cite{sobel1983accuracy} edge maps, and finally downsampled to the original resolution. Such a simple patch strategy could generate more precise BDCN~\cite{he2019bi} edge maps from the high-resolution input with less memory comsumption.
 % As illustrated in ~\cref{fig:edges_com},
 In Sec.~\ref{sec:ablation}, we compare our hybrid edge detection strategy with other methods, and the results indicate our fused edges could fit the accurate edges more closely compared to methods only relying on neural networks. Please see more results in the supplementary material.

\begin{table*}
\centering
    \caption{Quantitative experimental results. \textbf{Bold} figures indicate the best and \underline{underlined} figures indicate the second best. We test our method on three public datasets and select three \textit{SOTA} MDE models~\cite{Ranftl2021,Ranftl2022,yin2021learning} to test the effectiveness of our method (we choose two CNN-based MDE method and employ LeRes~\cite{yin2021learning} in DIODE~\cite{vasiljevic2019diode}). We also show the experimental results of the ablation studies in this table. Note that $F_{BC}$ is equal to $GF (B+C)$ which denotes fusing $B$ and $C$ using guided filters~\cite{he2010guided}, and $L_{BC}$ is equal to $LFM (B+C)$ which means fusing $B$ and $C$ with LFM. The methods of (d)-(i) do not use the DCM while the methods of (k)-(n) use it. \textit{Ours-BDCN} and \textit{Ours-Sobel} indicate our full method with only BDCN edges~\cite{he2019bi} and Sobel edges~\cite{sobel1983accuracy}.}

    \renewcommand{\arraystretch}{0.9}
    \resizebox{\linewidth}{!}{
    
    \begin{tabular}{lcccccccccccccc}
    \toprule
    \multirow{2}{*}{Method} & \multicolumn{4}{c}{DIODE \cite{vasiljevic2019diode}}  & \multicolumn{4}{c}{IBims-1 \cite{ibim2018evaluation}} & \multicolumn{4}{c}{NYU-v2 \cite{nyu2012indoor}} \\
    \cmidrule(lr){2-5} \cmidrule(lr){6-9} \cmidrule(lr){10-13}
    & SqRel$\downarrow$ & ESR$\downarrow$ & EcSR$\downarrow$ & ORD$\downarrow$ & SqRel$\downarrow$ & ESR$\downarrow$ & EcSR$\downarrow$ & ORD$\downarrow$ & SqRel$\downarrow$ & ESR$\downarrow$ & EcSR$\downarrow$ & ORD$\downarrow$\\
    \midrule
    
    (a) CNN-based \cite{Ranftl2022,yin2021learning} & 7.289 & 7.173 & 7.211 & 0.412 & 0.653 & 0.664 & 0.661 & 0.364 & 0.470 & 0.489 & 0.479 & 0.306 \\
    \midrule
    (b) GDF~\cite{dai2022multi} & 7.304 & 7.372 & 7.402 & 0.423 & 0.625 & 0.645 & 0.651 & 0.377 & 0.432 & 0.440 & 0.438 & 0.295 \\
    
    (c) BMD~\cite{Miangoleh2021Boosting}  & 7.450 & 7.631 & 7.696 & 0.443 & 0.642 & 0.669 & 0.672 & 0.387 & 0.489 & 0.506 & 0.499 & 0.311 \\
    \midrule
    (d) Ours (Edge$(B)$) & 7.275 & 7.370 & 7.355 & 0.420 & 0.895 & 0.953 & 0.942 & 0.424 & 0.635 & 0.651 & 0.654 & 0.421 \\
    
    (e) Ours (Edge-highlighted $(C)$)  & \textbf{7.206} & 7.126 & 7.178 & 0.402 & 0.666 & 0.660 & 0.655 & 0.358 & 0.480 & 0.489 & 0.490 & 0.352 \\
    
    (f) Ours (GF$(A+C)$)   & 7.321 & 7.419 & 7.401 & 0.419 & 0.656 & 0.646 & 0.653 & 0.366 & 0.454 & 0.483 & 0.478 & 0.299 \\
    
    (g) Ours (GF$(A+B)$)    & 7.314 & 7.409 & 7.399 & 0.424 & 0.693 & 0.710 & 0.706 & 0.371 & 0.477 & 0.485 & 0.488 & 0.349 \\
    
    (h) Ours (GF$(B+C)$) & 7.273 & 7.198 & 7.204 & 0.427 & 0.703 & 0.721 & 0.715 & 0.401 & 0.463 & 0.471 & 0.468 & 0.325 \\

    (i) Ours (GF$(A + F_{BC})$) & 7.342 & 7.405 & 7.399 & 0.430 & 0.674 & 0.683 & 0.671 & 0.372 & 0.470 & 0.487 & 0.479 & 0.309 \\
    \midrule
    (j) Ours-w/o DCM (LFM$(A+F_{BC})$) & 7.354 & 7.371 & 7.345 & 0.432 & 0.679 & 0.683 & 0.677 & 0.379 & 0.466 & 0.457 & 0.459 & 0.334 \\
    
    (k) Ours-BDCN (LFM$(A+F_{BC})$) & 7.313 & 7.222 & 7.209 & 0.409 & 0.541 & 0.540 & 0.548 & 0.349 & \underline{0.402} & \underline{0.396} & 0.394 & 0.275 \\
    
    (l) Ours-Sobel (LFM$(A+F_{BC})$) & 7.276 & \underline{7.016} & \underline{7.071} & \underline{0.399} & \textbf{0.533} & \underline{0.521} & \underline{0.517} & \underline{0.333} & 0.418 & 0.398 & \underline{0.392} & \underline{0.290} \\
    
    (m) Ours-Full (LFM$(A+L_{BC})$) & 7.252 & 7.373 & 7.415 & 0.407 & 0.693 & 0.689 & 0.679 & 0.389 & 0.456 & 0.464 & 0.459 & 0.336 \\
    
    (n) Ours-Full (LFM$(A+F_{BC})$) & \underline{7.236} & \textbf{6.997} & \textbf{7.035} & \textbf{0.394} & \underline{0.538} & \textbf{0.494} & \textbf{0.511} & \textbf{0.329} & \textbf{0.391} & \textbf{0.378} & \textbf{0.375} & \textbf{0.254} \\
    
    \midrule \midrule
    
    (a) Transformer-based \cite{Ranftl2021} & 7.629 & 7.572 & 7.597 & 0.451 & 0.680 & 0.670 & 0.666 & 0.386 & 0.533 & 0.554 & 0.552 & 0.347 \\
    \midrule
     (b) GDF~\cite{dai2022multi} & 7.522 & 7.542 & 7.551 & 0.449 & 0.613 & 0.620 & 0.624 & 0.375 & 0.531 & 0.532 & 0.535 & 0.335 \\
    
    (c) BMD~\cite{Miangoleh2021Boosting}  & 7.651 & 7.660 & 7.662 & 0.454 & 0.692 & 0.723 & 0.719 & 0.381 & 0.560 & 0.582 & 0.579 & 0.356 \\
    \midrule
    (d) Ours (Edge$(B)$) & 7.974 & 8.114 & 8.094 & 0.479 & 0.970 & 0.953 & 0.962 & 0.407 & 0.786 & 0.802 & 0.795 & 0.391 \\
    
    (e) Ours (Edge-highlighted $(C)$)  & \textbf{7.457} & 7.536 & 7.605 & 0.443 & 0.702 & 0.713 & 0.721 & 0.367 & 0.609 & 0.621 & 0.624 & 0.329 \\
    
    (f) Ours (GF$(A+C)$)   & 7.522 & 7.622 & 7.615 & 0.448 & 0.691 & 0.710 & 0.714 & 0.366 & 0.501 & 0.506 & 0.507 & 0.294 \\
    
    (g) Ours (GF$(A+B)$)    & 7.609 & 7.652 & 7.697 & 0.450 & 0.700 & 0.715 & 0.711 & 0.363 & 0.581 & 0.579 & 0.577 & 0.320 \\
    
    (h) Ours (GF$(B+C)$) & 7.514 & 7.641 & 7.711 & 0.443 & 0.718 & 0.728 & 0.721 & 0.376 & 0.514 & 0.520 & 0.522 & 0.329 \\

    (i) Ours (GF$(A + F_{BC})$) & 7.640 & 7.831 & 7.905 & 0.457 & 0.686 & 0.692 & 0.689 & 0.368 & 0.521 & 0.520 & 0.517 & 0.313 \\
    \midrule
    (j) Ours-w/o DCM (LFM$(A+F_{BC})$) & 7.556 & 7.560 & 7.588 & 0.446 & 0.692 & 0.690 & 0.688 & 0.369 & 0.598 & 0.594 & 0.592 & 0.304 \\
    
    (k) Ours-BDCN (LFM$(A+F_{BC})$) & 7.579 & 7.342 & 7.412 & 0.440 & 0.521 & 0.516 & 0.518 & 0.340 & \underline{0.490} & \underline{0.488} & \underline{0.485} & 0.288 \\
    
    (l) Ours-Sobel (LFM$(A+F_{BC})$) & 7.523 & \underline{7.260} & \underline{7.315} & \underline{0.438} & \textbf{0.518} & \underline{0.499} & \underline{0.502} & \textbf{0.335} & 0.501 & 0.496 & 0.495 & \underline{0.287} \\
    
    (m) Ours-Full (LFM$(A+L_{BC})$) & 7.537 & 7.570 & 7.575 & 0.455 & 0.623 & 0.620 & 0.622 & 0.372 & 0.549 & 0.551 & 0.553 & 0.319 \\
    
    (n) Ours-Full (LFM$(A+F_{BC})$) & \underline{7.463} & \textbf{7.202} & \textbf{7.224} & \textbf{0.436} & \underline{0.520} & \textbf{0.490} & \textbf{0.495} & \underline{0.337} & \textbf{0.489} & \textbf{0.468} & \textbf{0.470} & \textbf{0.280} \\
    
    \bottomrule
    \end{tabular}
    }
 
\label{Tab: Cmp with SOTA}
\end{table*}

\begin{figure*}[t]
	\centering

	\includegraphics[width=0.98\linewidth]{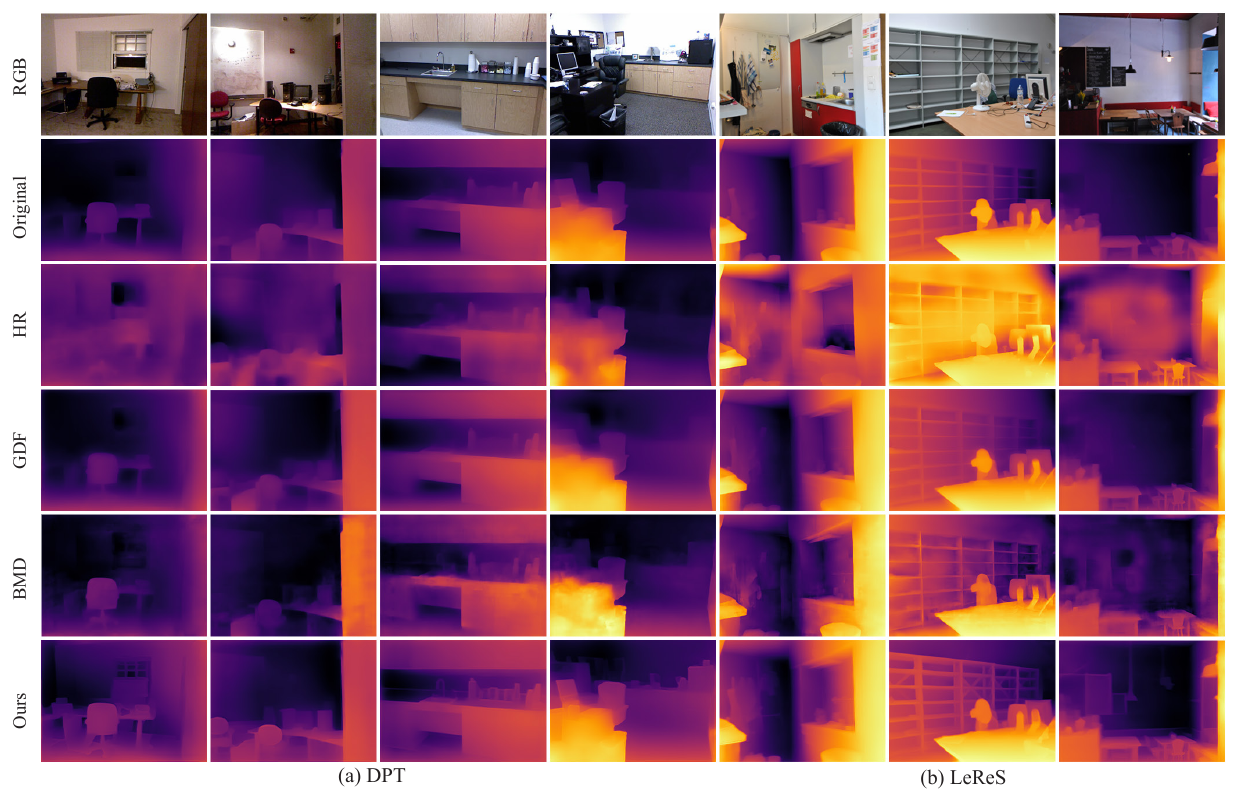}
    % \vspace{-1em}
	\caption{Visualization comparison with base models (DPT~\cite{Ranftl2021} and LeReS~\cite{yin2021learning}) and related fusion methods (BMD~\cite{Miangoleh2021Boosting} and GDF~\cite{dai2022multi}) on IBims-1~\cite{ibim2018evaluation} and NYU-v2~\cite{nyu2012indoor} datasets. \textit{Original} indicates the depth from base models, \textit{HR} indicates the depth of high-resolution inputs from base models. The left four columns use DPT~\cite{Ranftl2021} as the base MDE model, and the right three columns use LeRes~\cite{yin2021learning}.}
	\label{fig:visualization}
	% \vspace{-1em}
\end{figure*}

\subsection{Layered Fusion Module}\label{sec:method_lfm}
As mentioned in Sec.~\ref{sec:method_overview}, the initial depth $D$, $D_e$, and $D_{eh}$ possess complementary advantages: $D_e$ contains clear edges and details information but lacks accurate spatial structure, $D$ and $D_{eh}$ have better overall structure but their edges are not clear enough.
Such findings inspire us to fuse $D$, $D_e$, and $D_{eh}$ to leverage their respective strengths for producing high-quality depth maps with both clear edges and reasonable structure.

Specifically, we utilize guided filter~\cite{he2010guided} to fuse the different combinations of $D$, $D_e$, and $D_{eh}$ to find an optimal combination.
Through a comprehensive comparison (detailed results and analysis in Sec.~\ref{sec:ablation}), the combination of $(D+D_e+D_{eh})$ is optimal in terms of edge details and overall metrics.

To ensure the fused depth reserves high-frequency details, we propose a layered fusion module, named LFM, to perform the fusion of triad  process consists of two stages: $D_e$ and $D_{eh}$ are first fused, then $D$ is fused with the output of the first stage to produce the final result.

As we can use different fusion methods for different stages, we conduct adequate experiments in Sec.~\ref{sec:ablation} to determine the final design of LFM.
More specifically, we first use guided filter~\cite{he2010guided} to get a fused depth from $D_e$ and $D_{eh}$ as the guided filter reverses the high-frequency details of edge-related depth maps.
For the second fusion stage, we design a lightweight convolution network to fuse the depth obtained by guided filters with the original depth $D$.
This network uses two CNN-based branches to extract the features of input depths, then uses a shallow CNN-based decoder to fuse the features and output the fused depth. 

To train the aforementioned network, we follow the training paradigm proposed in BMD~\cite{Miangoleh2021Boosting} and GDF~\cite{dai2022multi}.
Specifically, we first use DPT~\cite{Ranftl2021} to infer the initial depths of the input RGB images with the size of 384×384 and 1024×1024, and then we resize all the initial depths to the size of 1024×1024 to generate training input pairs.
The pseudo label of each pair is obtained by using guided filters~\cite{he2010guided} to fuse the pairing depth.
% We use guided filters to fuse the input pairs as label data. 
Following SGR~\cite{Xian_2020_CVPR}, we use a gradient domain ranking loss to reserve more details, and use the same depth loss with LeRes~\cite{yin2021learning} to improve the scale consistency between the fused depth and the input depths.

% \subsection{Depth Consistency Module}
\subsection{Depth Consistency Module}\label{sec:method_dcm}
As analyzed in Sec.~\ref{sec:intro} and Sec.~\ref{sec:method_overview}, fusing multiple depth maps from different inputs, (\textit{i.e.}, original image, edge map, edge-highlighted image) is a stiff task, which is faced with two main challenges: (\romannumeral1) the depth of edge map contains clear edges but suffers from poor spatial structure, (\romannumeral2) the depth range and distribution of different depth inputs are inconsistent.
These challenges result in imperfect fused depth from LFM.
To handle this problem, we propose the DCM to fix the errors left by LFM and improve the depth consistency between the final depth and the input depths. Unlike previous consistent depth estimation methods~\cite{luo2020consistent,li2023towards}, we only unify the depth domain of the same scene, rather than predicting consistent depth for continuous motion with geometric constraints.
Specifically, the DCM takes as inputs the initial fused depth $D_{fuse}$ from LFM, the initial depth $D$ from original images $I$, and outputs the updated depth $D_{out}$, which can be represented as:
\begin{equation}
{D}_{out} ={D}_{fuse} +  DCM\left (   {D}_{fuse}, {D}   \right ).
\label{eq:dcm}
\end{equation}
According to Eq.~\ref{eq:dcm}, the DCM attempts to learn a depth residual to update the $D_{fuse}$ to minimize the inconsistency between $D_{fuse}$ and $D$.
To ensure the computing efficiency, we implement DCM with shallow convolution layers and residual blocks~\cite{he2016resnet}. The overall architecture of DCM consists of an encoder and a decoder with skip connections between them. 
% More details of DCM architecture could be found in supplementary materials.

Although the structure of DCM is concise, it is non-trivial to train DCM due to the lack of appropriate training data.
To this end, we develop a self-supervised training paradigm to train DCM.
Specifically, for each image $I$ in the HRWSI dataset~\cite{Xian_2020_CVPR}, we generate five images $\{I_s\}_{s=1}^{5}$ with different resolutions ranging from 384×384 to 1024×1024 as a group of training data.
Then we use the pre-trained fixed DPT~\cite{Ranftl2021} to produce initial depth maps corresponding to the aforementioned training images.
Before computing the loss, the initial depth maps will be resized and cropped into the same resolution as $\{D_s\}_{s=1}^{5}$.
As the high-resolution depth maps contain more high-frequency details, we use the same depth domain loss as MiDaS~\cite{Ranftl2022}, which constrains the similar depth domain, and a depth consistency loss as:

\begin{equation}
\label{eq:tc_loss_update}
D_{s}\Leftarrow D_{s} + DCM(D_{s}, D_{s-1}),
\end{equation}
\begin{equation}
\label{eq:tc_loss}
\mathcal L_{C}=\sum_{s=2}^{5} \left\|M_{s}\left(D_{s}-\hat{D}_{s-1}\right)\right\|_1,
\end{equation}
% 备份
% \begin{equation}
% \label{eq:tc_loss}
% \mathcal L_{C}=\sum_{s=2}^{5} \left\|M_{s}\left(D_{s}-\hat{D}_{s-1}\right)\right\|_1,
% \end{equation}

% \begin{equation}
% \label{tc_loss}
% \mathcal L_{C}=\frac{1}{|M_{s}|}\sum_{s=2}^{5} \sum_{p\in M_{s}}\left\|D_{s}(p)-\hat{D}_{s-1}(p)\right\|_{1},
% \end{equation}
where $M_{s}=\exp (-\alpha|V_{s}-\hat{V}_{s-1}|^{2})$ represents the occlusion weight between $D_{s}$ and $\hat{D}_{s-1}$, $\alpha$ is set to $50$.
$\hat{D}_{s-1}$ is obtained by warping the $D_{s-1}$ to $D_{s}$ according to the \textit{depth color} optical flow $F_{s \Rightarrow s-1}$ between $V_{s}$ and $V_{s-1}$, where the $V_{s}$ and $V_{s-1}$ are visualized images corresponding to $D_{s}$ and $D_{s-1}$.
% $D_{s-1}$ warped by the optical flow $F_{t \Rightarrow t-1}$, $V_{t}$ is the visualization color map of $D_{t}$, and
$\hat{V}_{s-1}$ is obtained by warping ${V}_{s-1}$ to $V_{s}$ according to the $F_{s \Rightarrow s-1}$. We use the FlowNet2~\cite{ilg2017flownet} to compute the backward flow $F_{s \Rightarrow s-1}$ between $V_{s}$ and $V_{s-1}$.
% $D_{s-1}$ warped by the optical flow $F_{t \Rightarrow t-1}$, $V_{t}$ is the visualization color map of $D_{t}$, and $\hat{V}_{t-1}$ is obtained by warping ${V}_{t-1}$ to $V_{t}$ according to the optical flow $F_{t \Rightarrow t-1}$. We use the FlowNet2~\cite{ilg2017flownet} to compute the backward flow $F_{t \Rightarrow t-1}$ between $V_{t}$ and $V_{t-1}$.

As shown in Fig.~\ref{fig:dcm}, we compare the visualized depth maps before and after DCM. Compared with the initial depth maps, the output depth maps possess significantly superior consistency.

\section{Experiments}\label{sec:exp}

\subsection{Datasets and Metrics}

We evaluate our method in three commonly used datasets, DIODE~\cite{vasiljevic2019diode}, IBims-1~\cite{ibim2018evaluation}, and NYU-v2~\cite{nyu2012indoor}. 
We adopt the SqRel, AbsRel and RMSE, as the standard evaluation metrics, and ordinal error (ORD)~\cite{Xian_2020_CVPR} to reveal finer details when evaluating with degraded images. We also use Edge Square Relative error (ESR and EcSR) to evaluate the edge depth quality. More details about these metrics are provided in the supplementary material. Besides, since depth suffers from scale ambiguity, we follow the standard depth evaluation protocol to align the predicted depth and the ground truth depth using the least squares method.

\begin{table}[t]
\renewcommand{\arraystretch}{1.2}
\setlength{\tabcolsep}{3pt}
\caption{Quantitative comparisons of the degraded images of IBims1-Noise~\cite{ibim2018evaluation} dataset. 
\label{Tab: Degraded}
}
\resizebox{\linewidth}{!}{%
\begin{tabular}{l|ll|ll|ll|ll}
\toprule[1pt]

\multirow{3}{*}{Method}  & \multicolumn{4}{c|}{IBims1-Noise-gaussian~\cite{ibim2018evaluation}} & \multicolumn{4}{c}{IBims1-Blur-gaussian~\cite{ibim2018evaluation}} \\ \cline{2-9}

&\multicolumn{2}{c|}{DPT~\cite{Ranftl2021}} & \multicolumn{2}{c|}{MiDaS~\cite{Ranftl2022}} & \multicolumn{2}{c|}{DPT~\cite{Ranftl2021}} & \multicolumn{2}{c}{MiDaS~\cite{Ranftl2022}} \\  

&\multicolumn{1}{c}{SqRel$\downarrow$} & \multicolumn{1}{c|}{ESR$\downarrow$ } & \multicolumn{1}{c}{SqRel$\downarrow$} & \multicolumn{1}{c|}{ESR$\downarrow$} & \multicolumn{1}{c}{SqRel$\downarrow$} & \multicolumn{1}{c|}{ESR$\downarrow$ } & \multicolumn{1}{c}{SqRel$\downarrow$} & \multicolumn{1}{c}{ESR$\downarrow$ } \\ \hline

Base Model  & \multicolumn{1}{c}{$0.8116$} &\multicolumn{1}{c|}{$0.8753$} &\multicolumn{1}{c}{$0.7729$} &\multicolumn{1}{c|}{$0.8531$} &\multicolumn{1}{c}{$0.7412$} & \multicolumn{1}{c|}{$0.8124$} &  \multicolumn{1}{c}{$0.7159$} &\multicolumn{1}{c}{$0.7412$}  \\ 

GDF~\cite{dai2022multi} &\multicolumn{1}{c}{$0.7744$}  &\multicolumn{1}{c|}{$0.9012$}  &\multicolumn{1}{c}{$0.7160$}& \multicolumn{1}{c|}{$0.7943$} &\multicolumn{1}{c}{$0.7196$} &\multicolumn{1}{c|}{$0.7743$} &\multicolumn{1}{c}{$0.7602$} &\multicolumn{1}{c}{$0.8025$} \\
BMD~\cite{Miangoleh2021Boosting} &\multicolumn{1}{c}{$0.8344$} &\multicolumn{1}{c|}{$0.8719$} &\multicolumn{1}{c}{$0.8032$} &\multicolumn{1}{c|}{$0.8952$} & \multicolumn{1}{c}{$0.7602$}  & \multicolumn{1}{c|}{$0.8432$ }&\multicolumn{1}{c}{$0.7421$} &\multicolumn{1}{c}{$0.7922$}    \\

\hline

Ours-Sobel  &\multicolumn{1}{c}{$\underline{0.6836}$} &\multicolumn{1}{c|}{$\underline{0.6692}$}  &\multicolumn{1}{c}{$\textbf{0.6524}$ } &\multicolumn{1}{c|}{$\underline{0.6371}$ } &\multicolumn{1}{c}{$\underline{0.6837}$} &\multicolumn{1}{c|}{$\underline{0.6654}$} &\multicolumn{1}{c}{$\underline{0.6203}$} &\multicolumn{1}{c}{$\underline{0.6019}$ } \\  

Ours &\multicolumn{1}{c}{$\textbf{0.6774}$ }&\multicolumn{1}{c|}{$\textbf{0.6431}$}  &\multicolumn{1}{c}{$\underline{0.6607}$}  &\multicolumn{1}{c|}{$\textbf{0.6311}$ } &\multicolumn{1}{c}{$\textbf{0.6021}$ }&\multicolumn{1}{c|}{$\textbf{0.5947}$ }& \multicolumn{1}{c}{$\textbf{0.6118}$}  &\multicolumn{1}{c}{$\textbf{0.5835}$ }

\\ \toprule[1pt]
\end{tabular}}
\vspace{-1em}

% \vspace{-1.5em}
\end{table}

% \subsection{Experimental Results}
\subsection{Experiments on Public Datasets}
To evaluate our proposed method, we test and compare ECFNet with other related methods using the aforementioned datasets.
In Tab.~\ref{Tab: Cmp with SOTA}, we use DPT~\cite{Ranftl2021}, LeRes~\cite{yin2021learning}, and MiDaS~\cite{Ranftl2022} as base MDE models, and compare ECFNet with base models and another two \textit{SOTA} depth fusion methods, (\textit{i.e.}, BMD~\cite{Miangoleh2021Boosting} and GDF~\cite{dai2022multi}).
According to the quantitative results in (a)-(c) and (n) of Tab.~\ref{Tab: Cmp with SOTA}, ECFNet significantly improves the overall quality of initial depth from the base models, especially in NYU-v2 dataset, where our method achieves around $8\%$ and $17\%$ improvement compared to  DPT~\cite{Ranftl2021} and MiDaS~\cite{Ranftl2022} respectively in terms of the SqRel metric.
When compared with \textit{SOTA} fusion methods (BMD~\cite{Miangoleh2021Boosting} and GDF~\cite{dai2022multi}), ECFNet also obtains a remarkable performance gap.

\begin{figure}[t]
  \centering
  \includegraphics[width=0.47\textwidth]{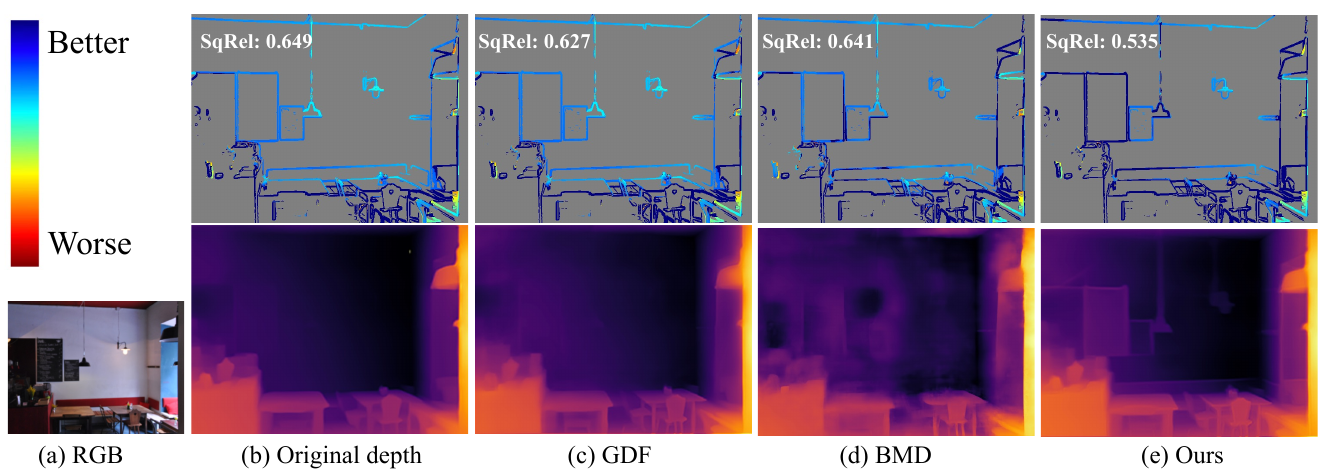}

\vspace{-1em}
  \caption{ Depth errors visualizations. 
  }
  
  \label{fig:depth edge error}
  % \vspace{-1em}
\end{figure}

In Fig.~\ref{fig:visualization}, we compare the visualized depth results of base MDE models (DPT~\cite{Ranftl2021} and LeRes~\cite{yin2021learning}) and related depth fusion methods (BMD~\cite{Miangoleh2021Boosting} and GDF~\cite{dai2022multi}) on IBims-1~\cite{ibim2018evaluation} and NYU-v2~\cite{nyu2012indoor} datasets. The predicted depth of ECFNet contains much clearer edges and better overall structure than depth maps from other methods.
Following the common practice, we test the base models with high-resolution input images for better details, but the results contain serious artifacts and fewer details than our results. We also show edge depth error maps in Fig.~\ref{fig:depth edge error} to validate that we can effectively improve the edge depth accuracy.

To further evaluate the trade-off between the inference efficiency and the quality of the depth map produced by our method, we increase the input resolution of base models while maintaining the same runtime of approximately 100ms. As shown in Tab. ~\ref{tab:runtime}, we present compelling evidence that our method outperforms the baseline model significantly under equal runtime conditions.

\subsection{Experiments on Degraded Images}

\begin{figure}[t]

  \begin{center}
	\includegraphics[width=0.95\linewidth]{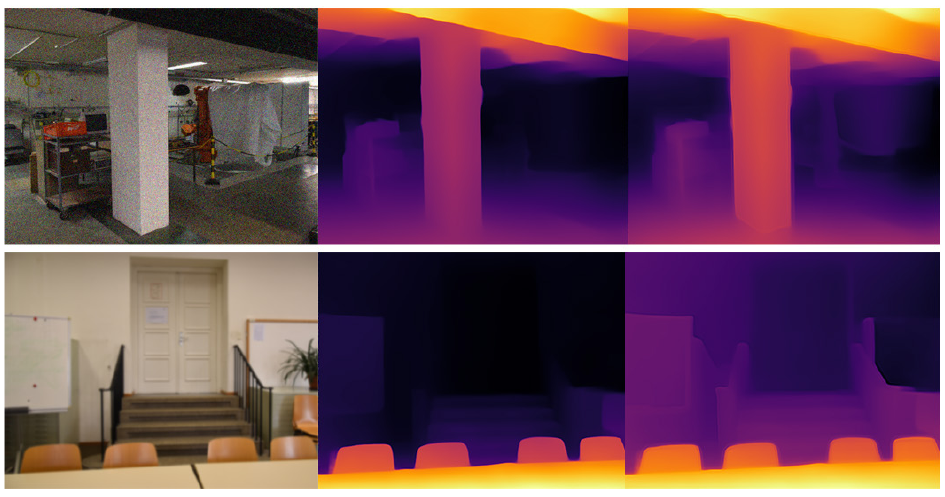}
	\caption{Depth visualization on degraded images of IBims1~\cite{ibim2018evaluation} dataset. Each triplet consists of the degraded RGB image, the corresponding depth predicted by DPT~\cite{Ranftl2021} and our method.}
	\label{fig:degrad_vis}
  \end{center}
    \vspace{-1.5em}
\end{figure} 

In this subsection we test ECFNet with the degraded images to evaluate its robustness. Concretely we compare ECFNet with other \textit{SOTA} methods on \textit{IBims1-noise-gaussian-001}  and \textit{IBims1-blur-gaussian-1-7783} data sequences~\cite{ibim2018evaluation}, where images are corrupted by the Gaussian noise and Gaussian blur, respectively. 

Tab.~\ref{Tab: Degraded} reports the quantitative results. ECFNet still outperforms its counterparts under degraded datasets, and the performance of ECFNet on degraded images is even on par with that of original methods on normal images. Then we visualize the depth maps predicted from degraded images by the base model and ECFNet in Fig.~\ref{fig:degrad_vis}. It could be found that the depth maps of ECFNet do not deteriorate significantly with image degradation, but retain no artifacts in noisy images and keep clear in blur images. This is probably because ECFNet relies on edge maps rather than high-resolution input to enhance the high-frequency details, and thus is less likely to be affected by image degradation due to the robust edge strategy.

However, when faced with noisy and blur images, a straightforward solution is to adopt the restoration algorithms as the image preprocessing strategies. Therefore, we further compare ECFNet with the solution above. As shown in Tab.~\ref{Tab:Res}, ECFNet surpasses these restoration methods under all six metrics even without specific designs for the degradation. As a result, ECFNet turns out to be a novel method to estimate depth maps with high-frequency details on low-quality and high-noise images.

\begin{table}[!htb]
\scriptsize
\centering
\renewcommand{\arraystretch}{0.9}
\setlength{\tabcolsep}{5pt}

\caption{Quantitative comparison results of inference efficiency, where our method outperforms other methods in the same inference time.}
\label{tab:runtime}

\resizebox{\linewidth}{!}{%
\begin{tabular}{l|l|ll|ll}
\toprule[1pt]

\multirow{2}{*}{Method}  &\multicolumn{1}{c|}{Per-frame}
&\multicolumn{2}{c|}{IBims-1~\cite{Ranftl2021}} & \multicolumn{2}{c}{DIODE~\cite{Ranftl2022}} \\  

&Runtime & \multicolumn{1}{c}{SqRel$\downarrow$} & \multicolumn{1}{c|}{ESR$\downarrow$ } & \multicolumn{1}{c}{SqRel$\downarrow$} & \multicolumn{1}{c}{ESR$\downarrow$ } \\ \hline

DPT~\cite{Ranftl2021} & $\approx$31ms & $0.689$ & $0.702$ &  $7.702$ &$7.721$  \\ 
LeRes~\cite{he2010guided} & $\approx$31ms &$0.624$ &$0.630$ &$7.431$ &$7.474$  \\
DPT~\cite{Ranftl2021} &$\approx$100ms &$0.839$ &$0.844$ &  $7.991$ &$7.931$  \\ 
LeRes~\cite{he2010guided} & $\approx$100ms &$0.744$ &$0.765$ &$7.730$ &$7.776$  \\

GDF~\cite{dai2022multi} &$\approx$185ms &$0.612$ &$0.619$ &$7.522$ &$7.542$ \\
BMD~\cite{Miangoleh2021Boosting}& $\approx$14750ms& $0.691$  &$0.722$ &$7.651$ &$7.660$    \\

\hline

Ours-DPT  & $\approx$100ms&$\underline{0.519}$ &$\underline{0.489}$ &$\underline{7.463}$ &$\underline{7.202}$  \\  

Ours-LeRes & $\approx$100ms&$\textbf{0.452}$ &$\textbf{0.434}$ &$\textbf{7.235}$  &$\textbf{6.997}$ 

\\ \toprule[1pt]

\end{tabular}%
}

\end{table}

\begin{table}[!htb]
\scriptsize

\centering
\renewcommand{\arraystretch}{0.8}
\setlength{\tabcolsep}{3pt}

\label{Tab: restoration}
\caption{Quantitative comparison of our method with image restoration methods~\cite{Zamir2020MIRNet}.}

\resizebox{\linewidth}{!}{

\begin{tabular}{l|llllll}
\toprule[1pt]

\multirow{3}{*}{Method}  & \multicolumn{6}{c}{IBims1-noise-gaussian-001~\cite{ibim2018evaluation}}  \\ \cline{2-7}

&\multicolumn{1}{c}{AbsRel$\downarrow$} & \multicolumn{1}{c}{ESR$\downarrow$} & \multicolumn{1}{c}{RMSE$\downarrow$} & \multicolumn{1}{c}{$\delta_{1}\uparrow$ } & \multicolumn{1}{c}{ORD$\downarrow$} & \multicolumn{1}{c}{EcSR $\downarrow$ } \\ \hline

DPT~\cite{Ranftl2021}  & $0.156$ &$0.832$ &$3.352$ &$0.785$ &$0.551$ & $0.826$  \\ 
Denoising~\cite{Zamir2020MIRNet} & $0.152$ &$0.710$ &$3.170$ &$0.808$ &$0.553$ & $0.714$   \\ 
Ours &$\textbf{0.141}$  &$\textbf{0.664}$  &$\textbf{3.081}$& $\textbf{0.812}$ &$\textbf{0.548}$ &$\textbf{0.670}$ \\ \hline

\multirow{3}{*}{Method}  & \multicolumn{6}{c}{IBims1-blur-gaussian-1-7783~\cite{ibim2018evaluation}}  \\ \cline{2-7}

&\multicolumn{1}{c}{AbsRel$\downarrow$} & \multicolumn{1}{c}{ESR$\downarrow$} & \multicolumn{1}{c}{RMSE$\downarrow$} & \multicolumn{1}{c}{$\delta_{1}\uparrow$ } & \multicolumn{1}{c}{ORD$\downarrow$} & \multicolumn{1}{c}{EcSR  $\downarrow$ } \\ \hline

DPT~\cite{Ranftl2021}  &$0.148$ &$0.751$  &$3.228$  &$0.803$  &$0.613$ &$0.747$  \\ 
SR~\cite{Zamir2020MIRNet}  &$0.140$ &$0.702$  &$3.052$  &$0.808$  &$0.605$ &$0.699$ \\  
Ours &$\textbf{0.133}$ &$\textbf{0.589}$  &$\textbf{2.941}$  &$\textbf{0.814}$  &$\textbf{0.591}$ &$\textbf{0.594}$ 

\\ \toprule[1pt]

\end{tabular}
}

\label{Tab:Res}

\end{table}

\begin{figure}[t]
 % \vspace{-1em}
  \begin{center}
	\includegraphics[width=0.97\linewidth]{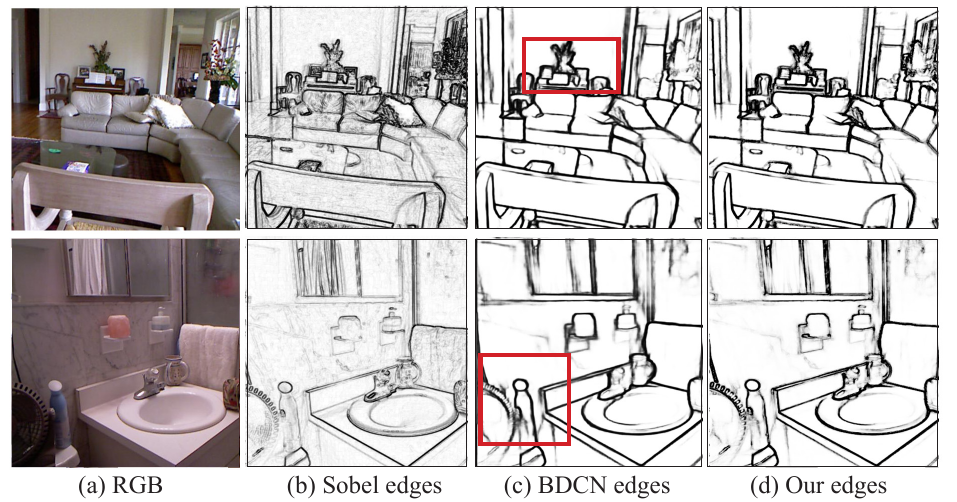}
 \vspace{-0.5em}
	\caption{Visualization comparison of different edges. Our edge maps contain fewer texture details and are closer to the real edge.}
	\label{fig:edges_com}
  \end{center}
   \vspace{-1em}
\end{figure} 

\begin{figure}[h]
  \centering
  \includegraphics[width=0.47\textwidth]{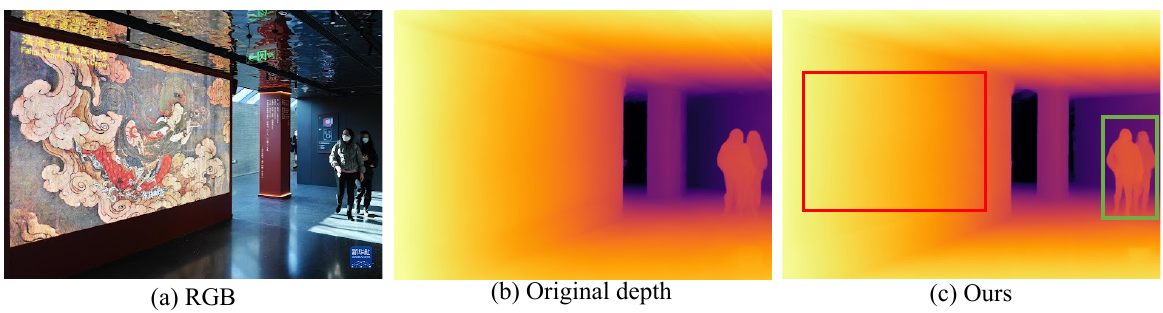}
  \vspace{-1em}
  \caption{ Depth map of the scene rich in textures. 
  }

\vspace{-1em}
  \label{fig:texture}
\end{figure}

\subsection{Ablation Studies}\label{sec:ablation}

\paragraph*{Hybrid edge detection strategy}
As ECFNet takes the edge map as input, the quality of the edge map would affect the final results.
In this section, we conduct ablation studies to validate the effectiveness of the hybrid edge detection strategy proposed in this paper. 

Concretely, we compare our hybrid edge detection strategy with the conventional edge detection algorithm Sobel operator~\cite{sobel1983accuracy} and learning-based edge detection method BDCN~\cite{he2019bi}. Firstly we focus on the quality of the edge.
As shown in Fig.~\ref{fig:edges_com}, Sobel edges contain undesired internal texture edges, while BDCN~\cite{he2019bi} edges are coarse in high-frequency regions. On the contrary, our edges are aligned with the actual object edges better. Then we further evaluate the effects of the three kinds of edges. According to (k), (l), and (n) of Tab.~\ref{Tab: Cmp with SOTA}, the hybrid edge detection strategy shows clear superiority over the other two methods. Hence we could conclude that the final results benefit from better edges, and we adopt the hybrid edge detection strategy in this work. 
As shown in Fig. ~\ref{fig:texture}, we effectively eliminate texture disturbances by our edge strategy, which avoids predicting unnecessary texture depth and maintaining model robustness without damaging the model’s priors.

\begin{figure}[t]
 % \vspace{-1em}
  \begin{center}
	\includegraphics[width=0.95\linewidth]{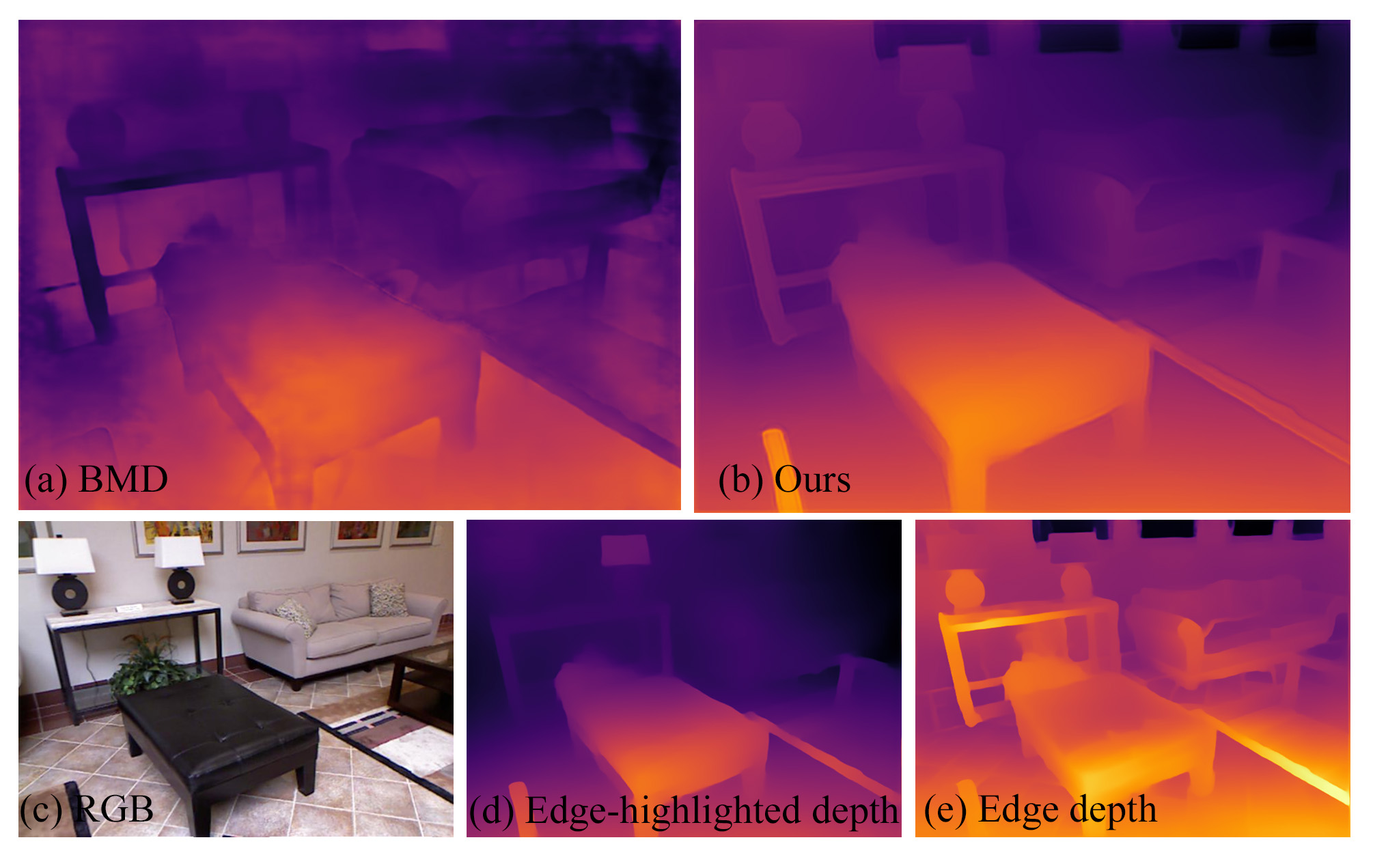}
	\caption{Visualization comparison of different fusion methods. Top (from left to right): depth of BMD~\cite{Miangoleh2021Boosting} and our LFM. Bottom (from left to fight): RGB, depth of edge-highlighted image, and edge map.}

	\label{fig:fusion}
  \end{center}
   \vspace{-1.5em}
\end{figure} 

\vspace{-0.5em}

\paragraph*{Candidate designs of LFM}
The target of LFM is to fuse the different depth maps of the original image, the corresponding edge map, and the edge-highlighted image. As introduced in Sec.~\ref{sec:method_lfm}, there are several different candidate designs for LFM.
In Tab.~\ref{Tab: Cmp with SOTA}, we test and compare the different schemes of LFM on four public datasets.
First, we compare the different combinations with 2 or 3 components in (f)-(i) of Tab.~\ref{Tab: Cmp with SOTA}.
In this group of experiments, we uniformly used the same guided filter (\textit{GF}) for all the methods, and the results indicate that utilizing all of the three depth components helps achieve the best performance.
Second, we test the different fusion manners in (i)-(j) and (m)-(n) of Tab.~\ref{Tab: Cmp with SOTA}.
The results of (i) and (j) show that using the network of LFM in the second fusion stage produces better results than using guided filter~\cite{he2010guided}, while the results of (m) and (n) indicate using the guided filter in the first fusion stage performs better. Thus we choose the method of (n) as our final scheme.
Fig.~\ref{fig:fusion} compares the LFM with the existing fusion method BMD~\cite{Miangoleh2021Boosting}, where the result shows LFM produces fewer artifacts and clear edges, implying the LFM is effective.

\vspace{-0.5em}
\paragraph*{Effectiveness of DCM}
DCM aims at minimizing the inconsistency between different depth inputs and producing high-quality depth with clear edges and reasonable structure.
In Tab.~\ref{Tab: Cmp with SOTA}, we test ECFNet with and without DCM using four public datasets.
According to the results of (j) and (n) of Tab.~\ref{Tab: Cmp with SOTA}, DCM plays an important role in improving the quality of final depth. Without DCM, though LFM could generate a clear depth edge, the final overall depth structure is undesirable. As a consequence, DCM is efficacious and indispensable.

\begin{figure}[t]
	\centering
	\includegraphics[width=0.99\linewidth]{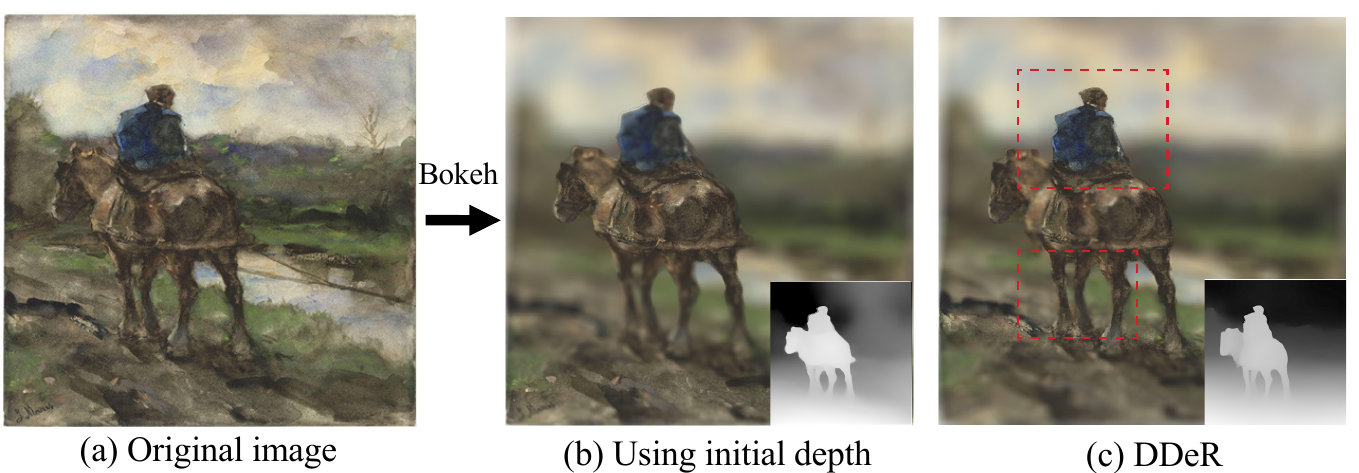}
	\caption{The predicted depth by ECFNet effectively enhances the edge performance of image applications.}
	\label{fig:edge_diffusion02}
	% \vspace{-0.5em}
 \vspace{-1em}
\end{figure}

\section{Potential applications}\label{appendix app}
Due to the diverse categories of artistic creations, existing depth estimation techniques struggle to accurately obtain depth maps for artistic images, which poses challenges for downstream tasks such as 3D photos~\cite{shih20203dphoto} and bokeh. However, our analysis shows that edges play a crucial role in depth estimation. By combining ControlNet~\cite{zhang2023adding} and diffusion models~\cite{rombach2022high} to generate cross-domain synthetic images, we can preserve the edge structure of the original artistic images and achieve cross-domain style transfer. As shown in Fig~\ref{fig:edge_diffusion02}, the resulting depth maps refined by our ECFNet have more details and clear edges by inputting the new synthetic image, leading to better performance in downstream tasks and promoting practical applications. We hope our findings and methods can drive the development of cross-domain image applications and inspire more edge-based approaches.

\section{Limitations}

Though ECFNet achieves great performance on various test datasets,
there are still several limitations as follows: 

\begin{enumerate}
\setlength{\itemsep}{0pt}
\setlength{\parsep}{0pt}
\setlength{\parskip}{0pt}
    \item[-] Compared to the base models or other lightweight MDE models, ECFNet requires more runtime and memory for inference.
    \item[-] The performance of ECFNet relies on the quality of the predicted edge maps, but there still remains room for improvement in the edge detection.
\end{enumerate}

\section{Conclussion}
In this paper, we propose ECFNet to estimate high-quality monocular depth with accurate edges and details using a single RGB image.
The overall pipeline of ECFNet is designed based on our exploration that the image edge information is critical to reserve high-frequency content.
In ECFNet, we first use the hybrid edge detection strategy to get the edge map and edge-highlighted image, which will be fed into a pre-trained MDE network together with the original RGB image to infer the initial depth.
Then we propose a layered fusion module (LFM) to fuse the aforementioned three initial depths, and the fused depth will further be optimized by our proposed depth consistency module (DCM) to form the final estimation.
As a result, ECFNet achieves significantly superior performance compared with other related methods on four public datasets. 
We hope that our approach will inspire more robust depth estimation algorithms and facilitate practical downstream applications.

\bibliographystyle{ieeenat_fullname}
\bibliography{sample-base}
% \clearpage
\appendix

\begin{figure*}[t]
	\centering
	\includegraphics[width=0.8\linewidth]{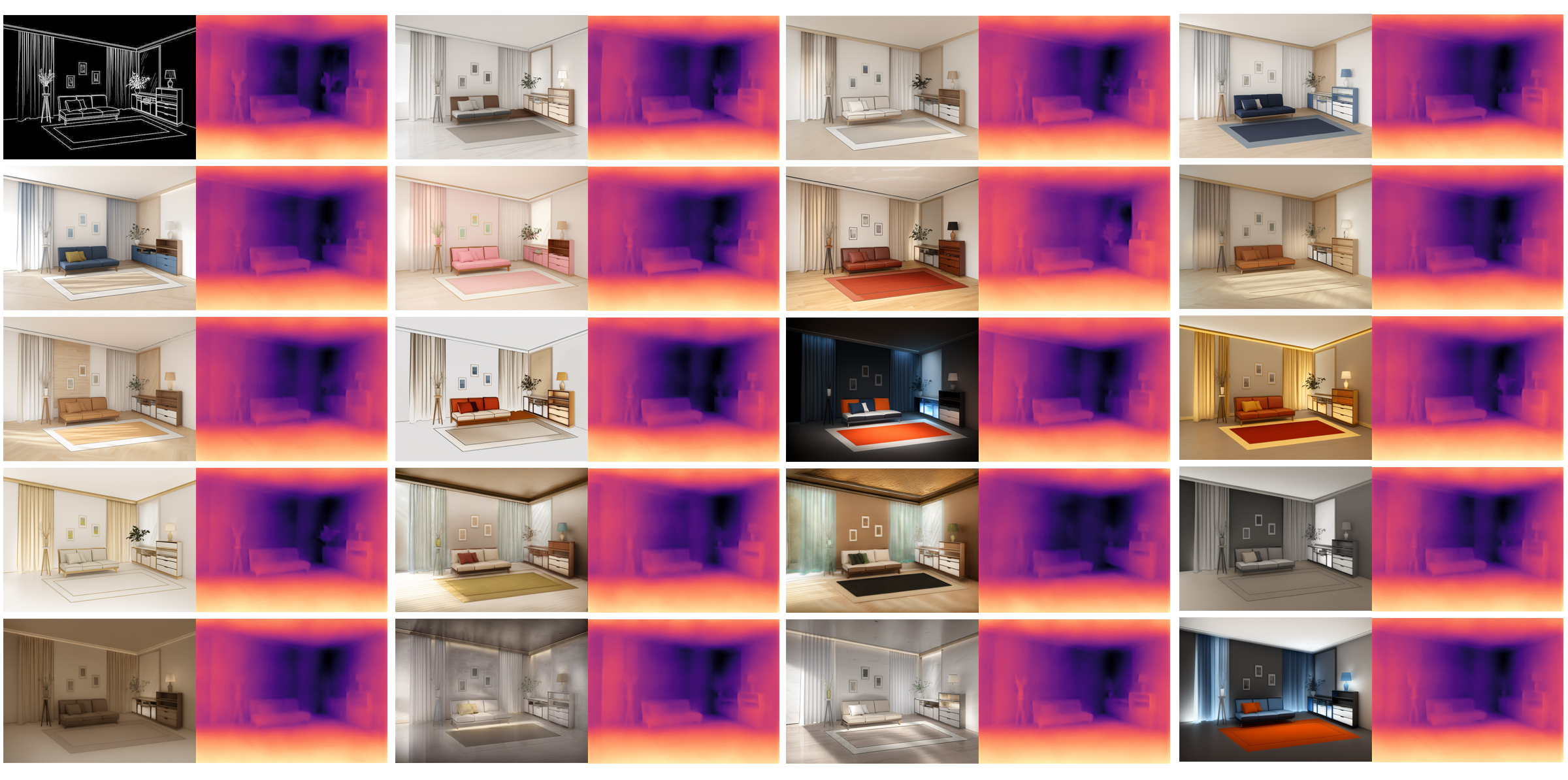}
	\caption{We present visualization results of depth maps predicted from various input images. Each image pair consists of an RGB image and the depth map predicted by DINOv2~\cite{oquab2023dinov2}. The structural information of these output depth maps is almost identical, as the corresponding input RGB images have very similar edge information.}
	\label{fig:edge_different}
\end{figure*}

% \begin{figure*}[h]
% 	\centering
% 	\includegraphics[width=0.99\linewidth]{./fig/diffusion_edge.jpg}
% 	\caption{We present visualization results of depth maps predicted from various input images. Each image pair consists of an RGB image and the depth map predicted by DINOv2~\cite{oquab2023dinov2}. The structural information of these output depth maps is almost identical, as the corresponding input RGB images have very similar edge information.}
% 	\label{fig:edge_different}
% 	% \vspace{-0.5em}
% \end{figure*}

\begin{figure*}[h]
	\centering
	\includegraphics[width=0.8\linewidth]{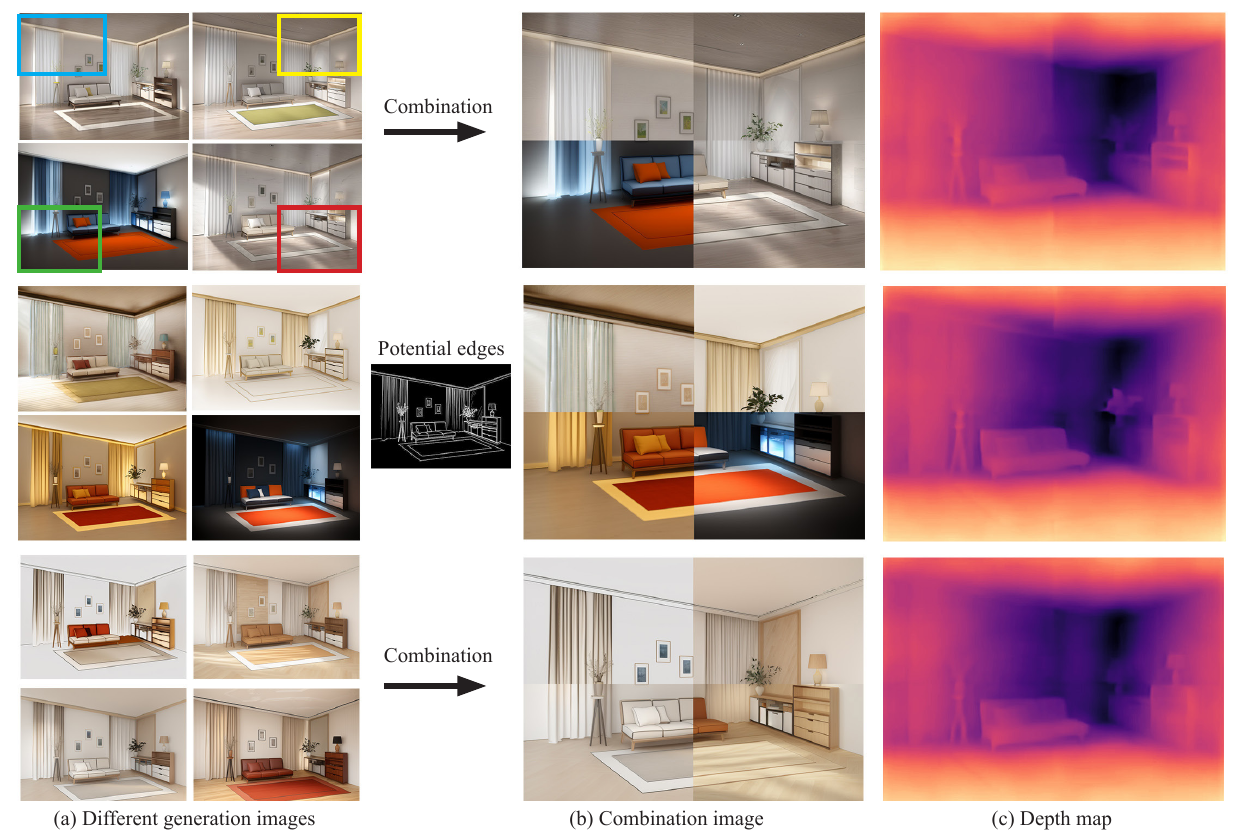}
	\caption{We form a group of four RGB images and extract 1/4 from each generation image to combine into a new image. The depth maps predicted from these combination images are almost identical because they share similar edge structure information.}
	\label{fig:edge_different01}
	% \vspace{-0.5em}
\end{figure*}

\begin{figure*}[h]
	\centering
	\includegraphics[width=0.9\linewidth]{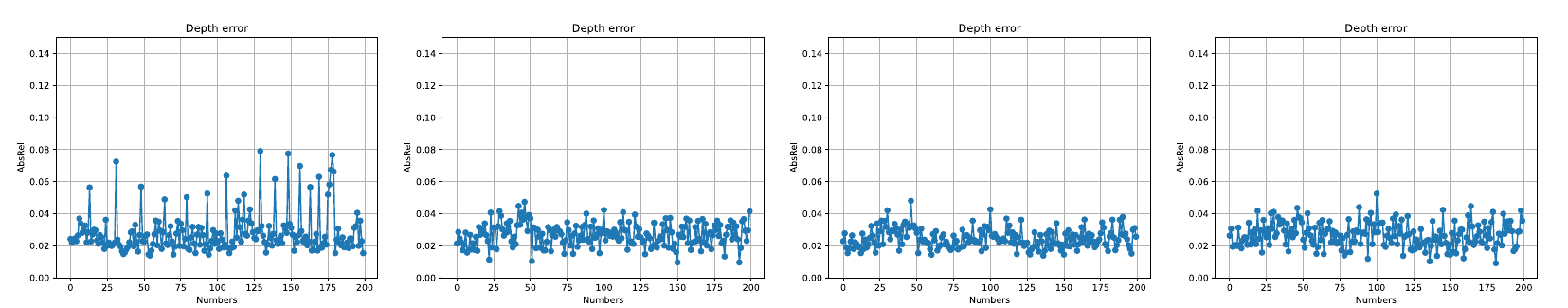}
	\caption{We show the absolute errors between each random pair of depth maps. The errors between these depth maps are negligible. Specifically, we show the unstable error results in the first figure to indicate that the diffusion model~\cite{rombach2022high} will generate some bad synthetic images in rare cases, thus making the corresponding depths also differ from the others.}
	\label{fig:same_depth_error}
   % \vspace{-5em}
\end{figure*}

\section{Appendix}
\subsection{Edge Information} \label{appendix edge}
In the main paper, we believe that edge information plays a crucial role in depth estimation. With the development of large-scale models~\cite{saharia2022imagen,ramesh2022hierarchical,rombach2022high} and edge-controlled image generation techniques~\cite{zhang2023adding}, it has become possible to generate images of the same scene with different styles but the same edge structure, which provides an opportunity to further validate the significance of edge information. To this end, we employ edge maps as structural control conditions and generate a large number of synthetic images using ControlNet~\cite{zhang2023adding} and stable diffusion models~\cite{rombach2022high}. These images have almost identical edge information but distinct textures and materials. Subsequently, we utilize the state-of-the-art depth estimation model DINOv2~\cite{oquab2023dinov2} to generate corresponding depth maps for each image, and remarkably, these depth maps exhibit highly consistent structural information, as demonstrated in Fig.~\ref{fig:edge_different}. To further verify that this phenomenon is attributed to the identical edge structure information, as shown in Fig.~\ref{fig:edge_different01}, we select 1/4 parts of 4 images (blue, yellow, green, red) from each group of images and combine them. Then, we predict the depth of the combined images. The generated depth maps still preserve high structural similarity.

To further quantify the difference among depth maps generated from these synthetic images, we leverage ControlNet~\cite{zhang2023adding} and stable diffusion models~\cite{rombach2022high} to generate 50 distinct synthetic images based on edge maps, and randomly arrange them to obtain 50 sets of depth maps. We compute the absolute error between each pair of depth maps, as illustrated in Fig.~\ref{fig:same_depth_error}, where we randomly sample several groups of results (approximately 200 groups of data per group) to demonstrate their absolute error quantification indicators. The errors between these depth maps are negligible due to their highly consistent edge structure information, which further corroborates the crucial role of edge information in depth estimation.

\subsection{Evalution Metrics}
\label{appendix metric}
We conduct our experiments in the depth space, and therefore the results predicted by MiDaS~\cite{Ranftl2022} and DPT~\cite{Ranftl2021} are processed by the formula $D = D_{max} - D$. We utilize the commonly applied depth estimation metrics, which are defined as follows:

\begin{itemize}[leftmargin=*]
\item Absolute relative error (AbsRel): $\frac{1}{n}\sum_{i=1}^n\frac{|d_i - d_i^*|_1}{d_i^*};$
\item Square relative error (SqRel): $\frac{1}{n}\sum_{i=1}^n\frac{(d_i - d_i^*)^2}{d_i^*};$
\item Edge Square relative error (ESR / EcSR): $ESR = \frac{1}{N}\sum_{i=1}^N E_{i}*\frac{(d_i - d_i^*)^2}{d_i^*}$,
$EcSR = \frac{1}{N}\sum_{i=1}^N Ec_{i}*\frac{(d_i - d_i^*)^2}{d_i^*}$
% which is defined as follows:
% \begin{equation}
% ESR = \frac{1}{N}\sum_{i=1}^N E_{i}*\frac{(d_i - d_i^*)^2}{d_i^*},
% \end{equation}
% dilated with a 5$\times$5 kernel
whert $E_{i}$ is the our edge binary mask pixel, $Ec_{i}$ means the canny edge~\cite{canny1986computational} binary mask pixel (the high and low thresholds are set to 100 and 200, respectively).  $N$ is the number of pixels.
\item Root mean squared error (RMSE): $\sqrt{\frac{1}{n}\sum_{i=1}^n(d_i - d_i^*)^2};$
\item Accuracy with threshold $t$: Percentage of $d_i$ such that $\\max(\frac{d_i}{d_i^*},\frac{d_i^*}{d_i}) = \delta_1<1.25;$
\item Ordinal error (ORD)~\cite{Xian_2020_CVPR}: 
$\frac{\sum_{i} \omega_{i} \mathbb{I}\left(\ell_{i} \neq \ell_{i, \tau}^{*}(p)\right)}{\sum_{i} \omega_{i}}$,
where $\omega_{i}$ is a weight to set to 1, ORD use the function with sample strategy between $\ell_{i}$ and $\ell_{i, \tau}^{*}(p)$. ORD is a general metric for evaluating the ordinal accuracy of a depth map.

\end{itemize}
where $n$ denotes the total number of pixels, $d_i$ and $d_i^*$ are estimated and ground truth depth of pixel $i$, respectively.

\paragraph{Loss Functions in Layer Fusion Module (LFM)}
For the loss function in the depth domain, we use the same loss as LeRes~\cite{yin2021learning}:
\begin{equation*}
\label{normalization loss}
  L_\text{ILNR} = \frac{1}{N}\sum_{i}^{N}\left|d_{i} - \overline{d}^{*}_{i}\right| + \left|tanh({d_{i}/100)} - tanh(\overline{d}^{*}_{i}/100) \right|,
\end{equation*}
where $\overline{d}^{*}_{i} =   ( d^{*}_{i} - \mu_\text{trim} ) $/${\sigma_\text{trim} }$ and $\mu_{\rm trim}$ and $\sigma_{\rm trim}$ are the mean and the standard deviation, we remove the maximum and minimum $10\%$ of depth values in the depth map. $d$ is the predicted depth, and $d^{*}$ is the ground truth depth map. 
Given a set of sampled point pairs $\mathcal{P}=\left\{\left[p_{i, 0}, p_{i, 1}\right], i=1, \ldots, N\right\}$, we use the same structure-guided ranking loss as SGR~\cite{Xian_2020_CVPR} in the gradient domain:
\begin{equation*}
\mathcal{L}_{\text {rank }}(\mathcal{P})=\frac{1}{N} \sum_{i} \phi\left(p_{i, 0}-p_{i, 1}\right),
\end{equation*}
More details about the sample strategy can be found in their paper~\cite{Xian_2020_CVPR}.

\subsection{Implementation Details}
\label{appendix details}
\begin{figure}[t]
	\centering
	\includegraphics[width=0.99\linewidth]{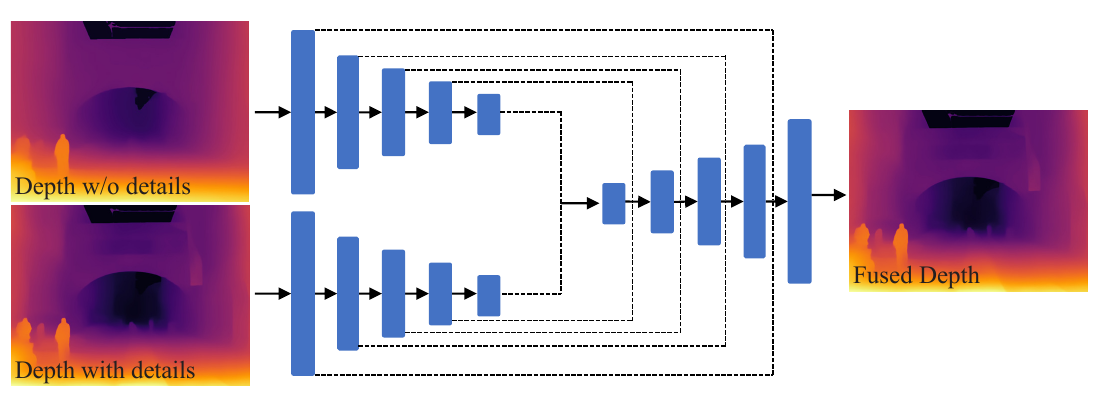}
	\caption{The network structure of our layer fusion module (LFM).}
	\label{fig:LFMNet1}
	\vspace{-0.5em}
\end{figure}

%% 我们首先使用大小
\paragraph*{Layer Fusion Module (LFM)} 
We utilize an encoder-decoder network with 10 layers to increase the training and inference resolution to 1024 $\times$1024. The network is similar to u-net~\cite{ronneberger2015unet}, to better capture the edge gradient information, we divide the input into two streams: one for the depth with more complete scene structure, and the other stream for the depth with more edge datails. We use a kernel of size 3 × 3. Group normalization~\cite{wu2018group} and leaky ReLUs~\cite{paszke2017leaky} are applied except the last layer. We show the network structure in Fig. ~\ref{fig:LFMNet1}. During training, we implement our model using PyTorch and train it with the AdamW solver~\cite{loshchilov2017adamw} for 9,000 iterations with a learning rate of 1e-4. To obtain our training data, we generate 3,000 training data pairs from the HRWSI~\cite{Xian_2020_CVPR} dataset using DPT~\cite{Ranftl2021}, where the low resolution is 384 × 384 and the high resolution is 1024 × 1024. Pseudo labeled data with high-frequency and less depth artifacts can be generated by the method of fusing them through the guided filter. The window radius of the guided filter is set to one-twelfth of the depth map width, and the edge threshold is 1e-12.

\begin{figure}[t]
	\centering
	\centering
	\includegraphics[width=0.99\linewidth]{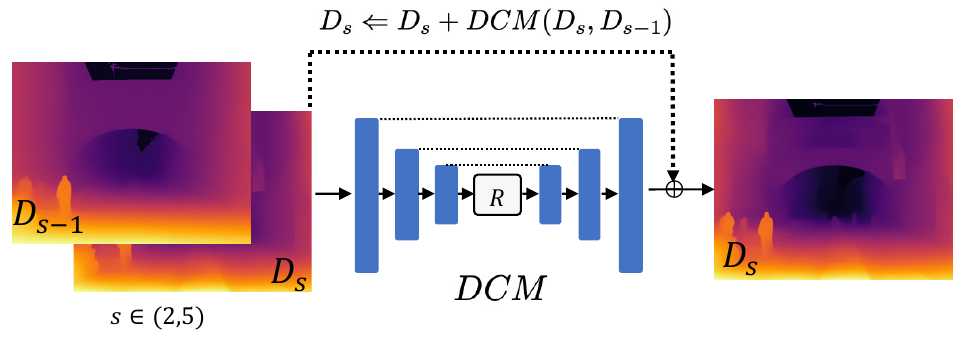}
	\caption{The network structure of our depth consistency module (DCM). $R$ represents the residual blocks.}
	\label{fig:DCMnet1}
\end{figure}

\paragraph*{Depth Consistency Module (DCM)}
The DCM utilizes an encoder-decoder architecture that is similar to Lai \textit{et al.}~\cite{Lai-ECCV-2018}. Specifically, the encoder consists of two downsampling strided convolutional layers, which are then followed by five residual blocks. 
The decoder consists of two transposed convolutional layers, and we incorporate skip connections from the encoder to the decoder. We show the network structure in Fig. ~\ref{fig:DCMnet1}.
Instance Normalization~\cite{ulyanov2017instance} is applied except the last layer. Finally, we apply a Tanh layer to ensure that the output is within the range from -1 to 1 after the decoder.
We implement our model using PyTorch and train it with the Adam solver~\cite{adam2015} for 50,000 iterations with a learning rate of 1e-4. During training, we use a batch size of 4 and randomly crop training data to 192×192.

\subsection{Edge strategy}
\label{appendix edge strategy}
BMD~\cite{Miangoleh2021Boosting} presents a comprehensive analysis of the network's receptive field size, focusing on its application to depth estimation. We can increase the resolution of the input image to obtain more precise depth estimation results. Interestingly, we also find that the resolution increase can benefit the generation of edge maps. Specifically, the higher resolution enables better localization of object boundaries, resulting in sharper and more detailed edges. This observation suggests that resolution enhancement can be a valuable approach for improving learning-based edge detection methods.

\begin{figure*}[h]
	\centering
	\includegraphics[width=0.9\linewidth]{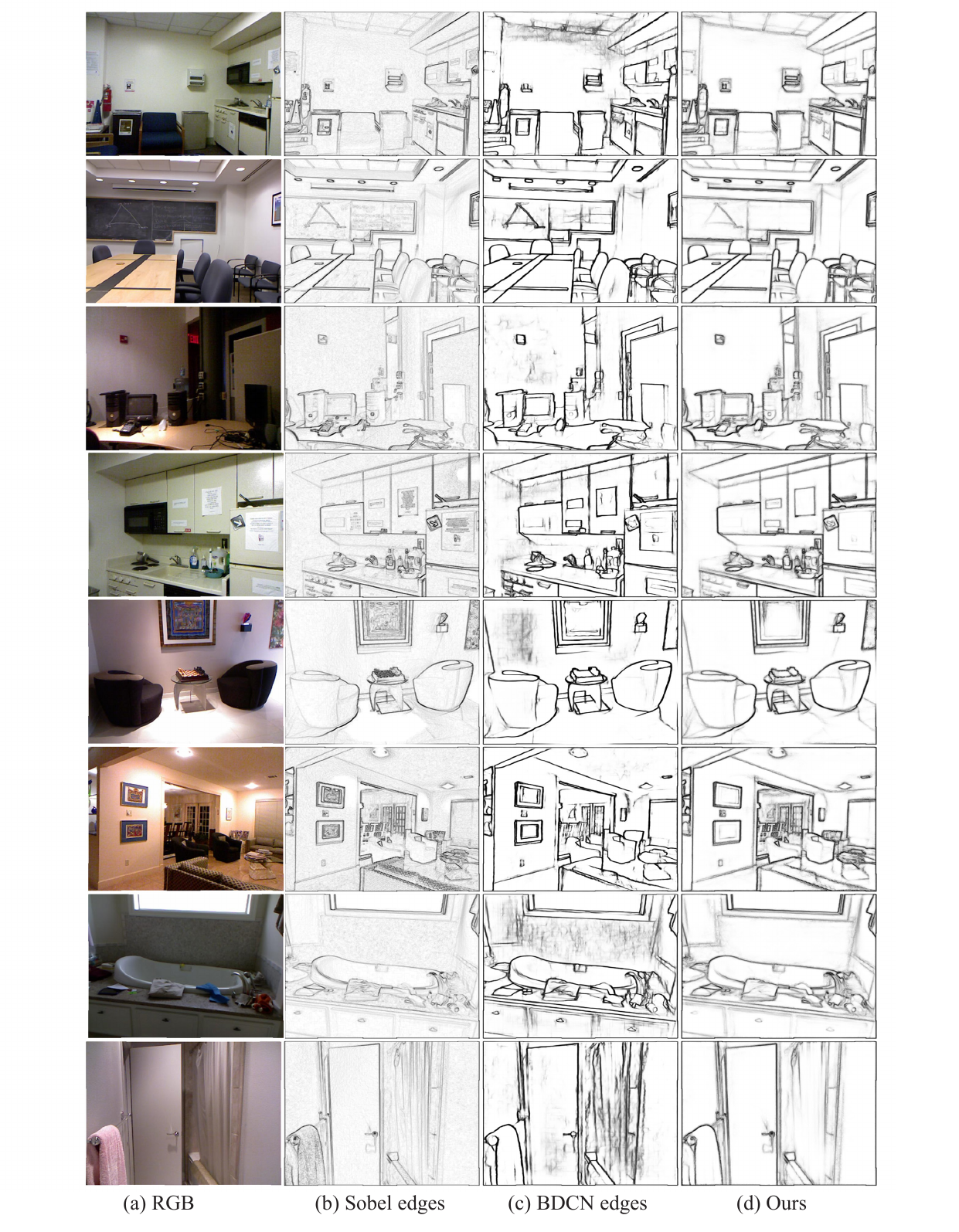}
	\caption{We show more visualization results of the edge map. The edge maps generated by our edge strategy are cleaner and closer to the actual edge maps.}
	\label{fig:edge_more}
	\vspace{-2em}
\end{figure*}

As a result, we appropriately increase the input resolution to enhance the accuracy of the estimated edge map,. Specifically, we first divide the image into nine uniform patches and upsample each patch to three times its original resolution. Then, these patches are fed into the BDCN network~\cite{he2019bi} to extract edge maps, which are subsequently fused with Sobel edge maps~\cite{sobel1983accuracy}. The fusion process is represented as $\sqrt[N]{E_{b} \times E_{s}}$, where $N$ is set to 2 to enhance contrast on both sides of the edge pixels, and $E_{b}$ and $E_{s}$ represent the BDCN edge and Sobel edge, respectively. This approach effectively eliminates hard edges and improves the overall smoothness and naturalness of the resulting image. Finally, we downsample the generated edge map to the original resolution. As shown in Fig. ~\ref{fig:edge_more}, we provide more edge maps generated by our edge strategy.

\subsection{Why use Guided Filters?}
\label{appendix gf}
\begin{figure}[t]
	\centering
	\includegraphics[width=0.99\linewidth]{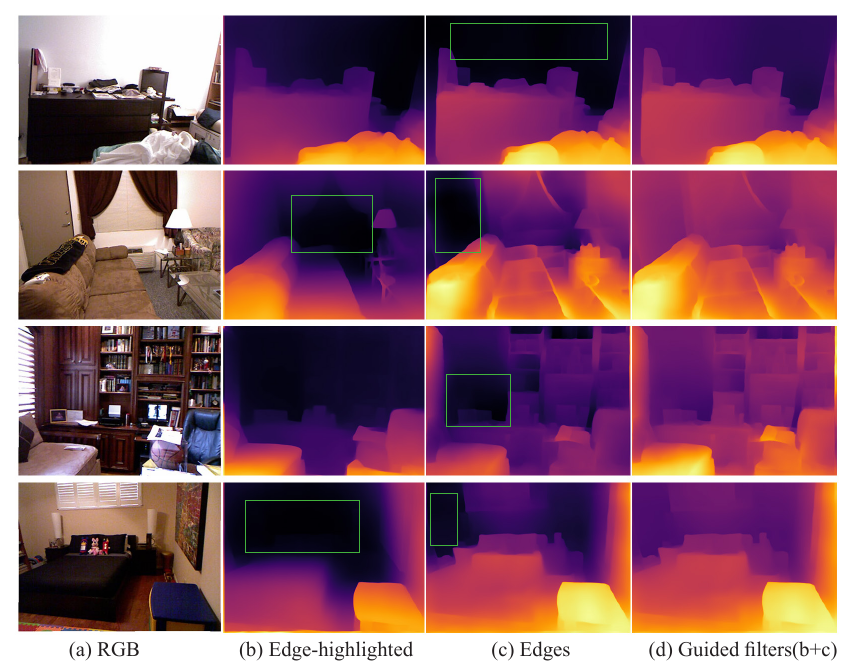}
	\caption{We show the fusion results of the guided filter, and the green boxes indicate the artifact areas in the depth map estimated from the edge map and the edge-highlighted image. The guided filter can effectively eliminate these artifacts.}
	\label{fig:gf}
	\vspace{-0.5em}
\end{figure}

The goal for the LFM is to seamlessly merge the complementary information coming from the depth maps which separately predicted by the original images, the edge maps and edge-highlighted images. As mentioned in the main paper, these depths possess complementary advantages. Actually, the depth map predicted by the edge map has the details we most expect to fuse, but it loses the overall structure of the scene and generates artifacts which cannot solve by directly using weighted averanging or Poisson blending. Guided filtering~\cite{he2010guided}can be an effective technique for our task, especially when the input depth maps have different scene structures and details that need to be preserved in the fused image.  We show the experimental comparison and visualization results of these two methods in Fig. ~\ref{fig:gf}.

\subsection{More Visualization Results}
\label{appendix vis}
\begin{figure*}[h]
	\centering
	\includegraphics[width=0.9\linewidth]{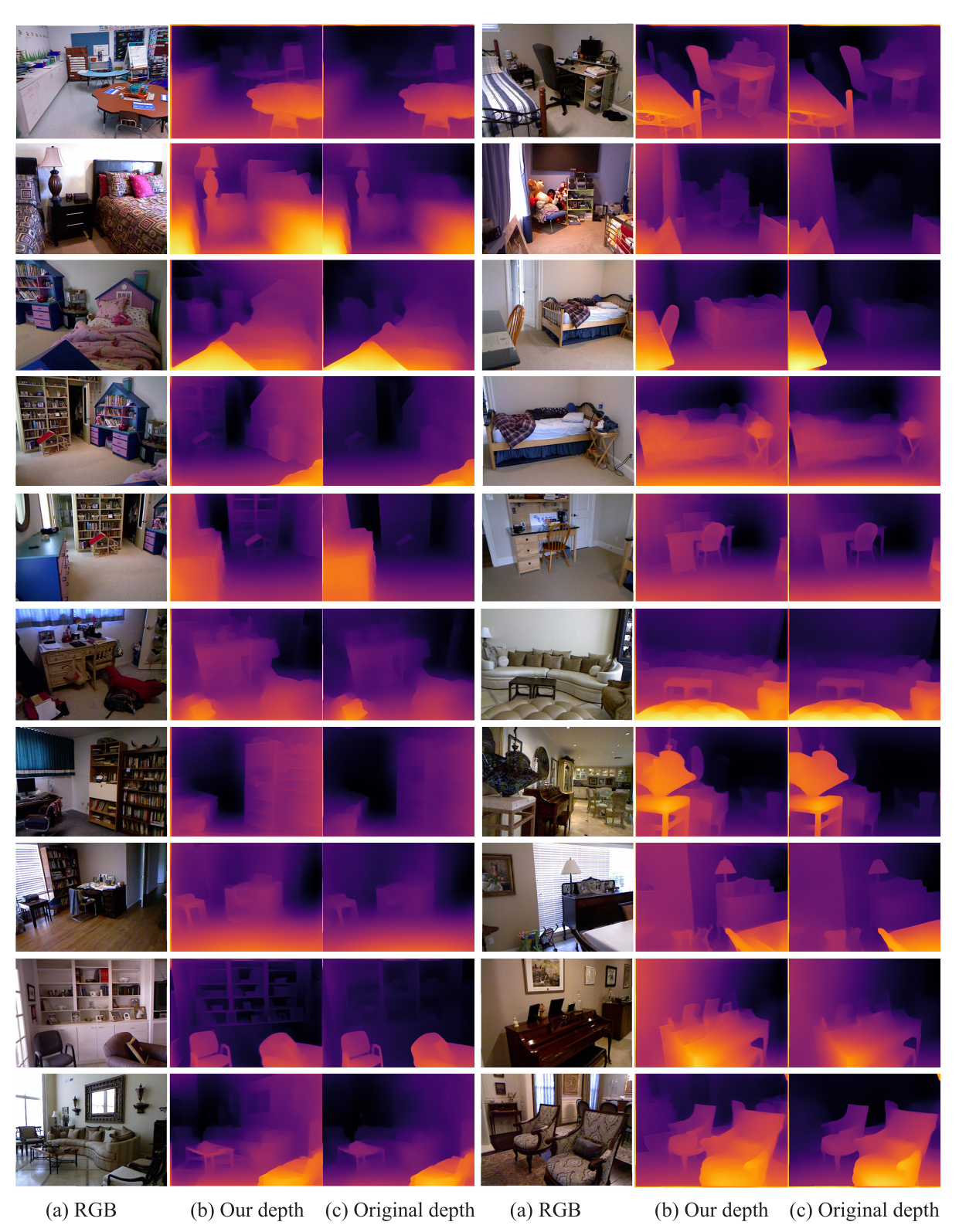}
	\caption{We provide more depth visualization. Each triplet consists of the RGB image, the depth predicted by our method, and the depth predicted by DPT~\cite{Ranftl2021}.}
	\label{fig:more_vis}
	\vspace{-2em}
\end{figure*}

We show more visualization results comparing the state-of-the-arts depth fusion methods with our method. As shown in Fig. ~\ref{fig:more_vis}, we achieve higher depth accuracy and obtains more edge details.

\end{document}